\documentclass[10pt,journal,compsoc]{IEEEtran}

\ifCLASSOPTIONcompsoc
  \usepackage[nocompress]{cite}
\else
  \usepackage{cite}
\fi
\usepackage{hyperref}
\usepackage{cite}
\usepackage{color}
\usepackage{amsmath}
\usepackage{graphicx}
\usepackage{float}
\usepackage{subfigure}
\usepackage{ragged2e}
\usepackage{blindtext}
\ifCLASSINFOpdf
\else
\fi
\begin{document}
%
\title{Constructing Predictive Surgical Path for AI-based Capsulorhexis Skill Transfer}
%
%
%
%
\author{Mohammad~Javad~Ahmadi
        and~Hamid~D.~Taghirad
\IEEEcompsocitemizethanks{
\IEEEcompsocthanksitem Both authors are with the Applied Robotics and AI Solutions (ARAS), Faculties of Electrical and Computer Engineering, K. N. Toosi University of Technology, Tehran, Iran.\protect\\
E-mail: mjahmadi@email.kntu.ac.ir, taghirad@kntu.ac.ir
}
}
\markboth{ }%
{Shell \MakeLowercase{\textit{et al.}}: Bare Advanced Demo of IEEEtran.cls for IEEE Computer Society Journals}
%



\IEEEtitleabstractindextext{%
\begin{abstract}
\justifying
Automated training of surgeons is one of the most crucial factors that significantly minimize surgical training risks and expenses.
With recent advances in artificial intelligence (AI) knowledge and available data from various surgeries, AI's involvement in surgical training is becoming very promising. It is recommended that at the early stages of AI development, it interferes in the surgery as a third agent alongside the trainer. As trust in AI increases, this process will lead to an AI agent acting as a trainer in the future. The first phase in which AI can intervene in the training process is to suggest an improved surgical path to the trainer. A platform must be constructed in the first step, to accomplish this task and to enhance the movement path of trainee surgeons. This paper introduces this platform along with an annotated capsulorhexis surgery dataset called the ARAS-Farabi (Cataract LMM) dataset. In this research, a deep convolutional neural network is pre-trained with JIGSAWS and ARAS-Farabi surgical datasets that can extract surgical skill characteristics from surgery tool tip motion data.  The proposed platform develops a reference model from the feature space of an expert surgeon's movement trajectory and proposes an improved path to enhance the skill of a novice surgeon. An optimization with two loss functions is utilized to create a path that raises the skill level of the novice surgeon's path while simultaneously predicting and preserving his/her intent. The results of this study reveal that, with the assistance of an AI agent, the trainee surgeon's movement path can be enhanced by at least 20 percent while maintaining his intentional objective. In addition to the recommended deep network, various tangible indicators have also been developed in this research to verify the level of trainee improvement.
\end{abstract}

}

\maketitle

\IEEEdisplaynontitleabstractindextext

%
\IEEEpeerreviewmaketitle

\ifCLASSOPTIONcompsoc
\IEEEraisesectionheading{\section{Introduction}\label{sec:introduction}}
\else
\section{Introduction}
\label{sec:introduction}
\fi
\IEEEPARstart{T}{he }
advances in robotics, control, and artificial intelligence (AI) facilitate the automation of the surgical training process, bringing ease and security to it. Several efforts have been made to automate surgical skill assessment. This is a big step toward automating surgical training; yet, these efforts need more advancements~\cite{c1,c2}. One of the most significant challenges in order to automate the surgical training process is finding an effective system for the transfer of surgical skills from expert surgeons to novices.

There are currently haptic and dual-user robotic devices that allow the trainer surgeon to instruct trainees on how to move and exert force during the surgery. These technologies may also be used to correct errors made by novice surgeons. There have been studies conducted on how to control these devices for surgical training; however, the majority of them lack the presence of AI agents. Consequently, the trainer surgeon is still investing a sizable amount of time and effort in the surgery training procedures.
Adding an AI agent in the surgical is very promising, in order to impart the master surgeon's surgical abilities to the trainee surgeon.

Our previous research focused separately on control structures and artificial intelligence to facilitate eye surgery training. The control structures used dual-user haptic systems and tracked errors between the trainer and trainee to transfer skills~\cite{c3,c4}. On the other hand, artificial intelligence automatically evaluated surgical performance by extracting skill features~\cite{c5}. To effectively train surgeons, a structure that combines control and artificial intelligence seems to be  essential. The first stage in this process is creating a platform that generates an expert path from the movements of a novice surgeon. In future works, this structure could serve as the interface between traditional control and artificial intelligence. It is also possible to integrate a predictive and educative artificial intelligence agent into robotic devices by incorporating skill-related knowledge obtained from artificial intelligence models that assess surgical skills.

This platform can be effectively used to enhance training circumstances, while eliminating the need for an experienced surgeon to supervise trainees, as well. Along with the trainee surgeon, this AI agent can first carry out the tasks of a third auxiliary agent (alongside a trained surgeon), before continuing to instruct surgery by itself.
Various initiatives have been undertaken in the past to facilitate the transfer of surgical skills. Ershad et al. developed an AI-enabled surgical training system that identifies the trainee's style-based movement flaws and produces haptic feedback to alert the trainee about existing errors \cite{c6}. However, this study did not provide the trainees with any guidance on how to enhance their actions. Zahedi et al. suggested an AI-based guidance system for a virtual kinesthetic training environment \cite{c61}. Fekri et al. created a platform for instructing new residents in orthopedic surgery drilling using a recurrent deep neural network \cite{c7}. Tan et al. developed a laparoscopic training robotic system that instructs students on how to manipulate surgical tools through demonstrations from human specialists as well as reinforcement learning \cite{c8}. The shortcoming of these studies is that they cannot account for new paths that the trainee might want to take during the surgery.

A key hindrance to the adoption of artificial intelligence (AI) in surgical training is the issue of trust. Constructing a foundational platform that highlights the ability of AI to enhance surgical practices while adhering to established surgical procedures, is an effective step in cultivating trust. Inspired by this requirement, this paper presents a dataset and AI-based system that identifies important information about an expert surgeon and injects those surgical skills into a trainee's behavior. In this study, the general features of surgery are extracted from the JIGSAWS dataset and injected into the motion path of novice surgeons. Additionally, this research resulted in the development of a dataset for capsulorhexis surgeries in order to focus on the expert surgical style and teach it to the trainees. 

The representation of the developed platform can be provided to the trainee surgeon in a visual format (Aras-Farabi surgical analysis software~\cite{c5}) or haptic feedback. The outline of this paper is as follows. The proposed procedure is described in \autoref{s1}. The intelligent approach, the suggested control structure, and the developed and utilized datasets are introduced in this section. The results are introduced and discussed in \autoref{s2}, while \autoref{s3} provides a summary, conclusions, and future directions.

\section{Methodology}\label{s1}
\begin{figure}[h!] 
\centering
\includegraphics[width=\columnwidth]{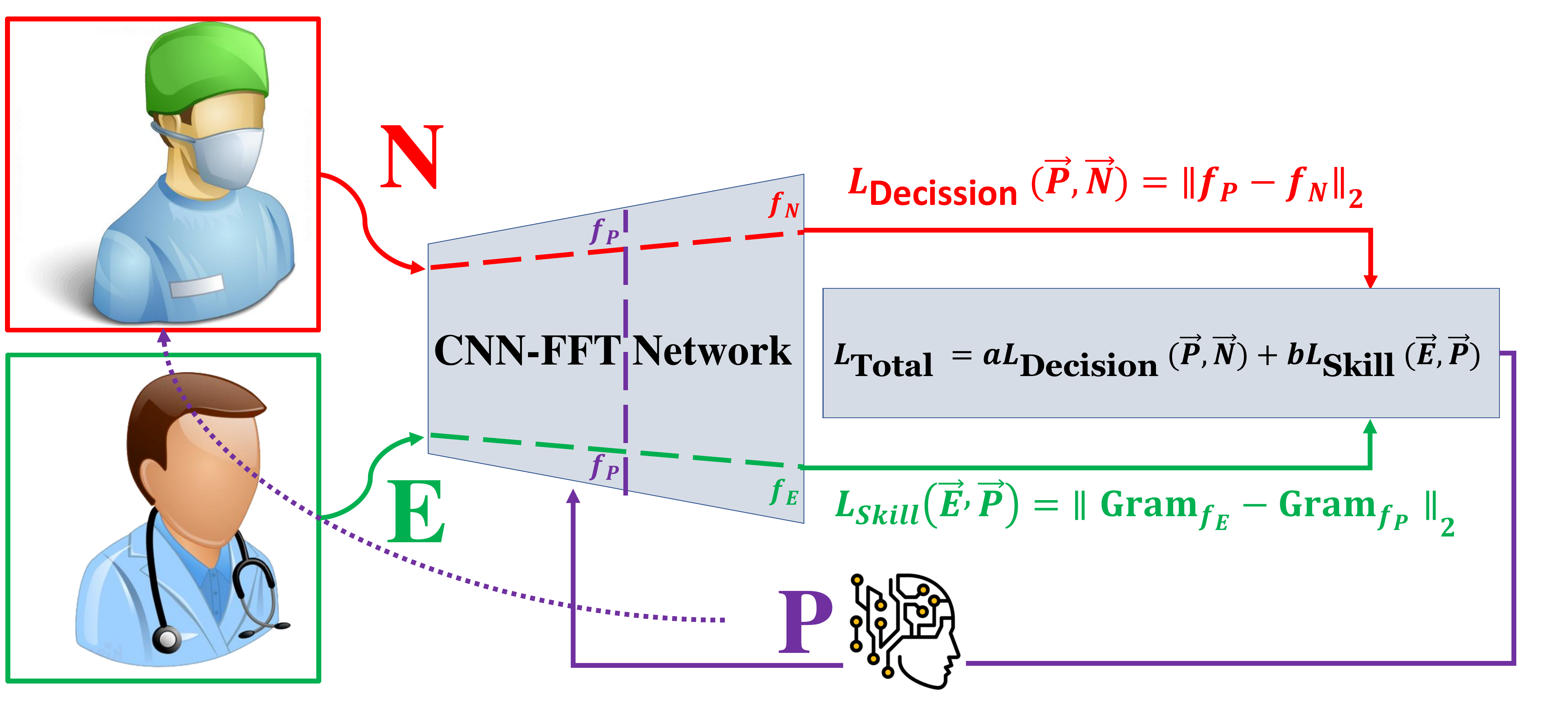}
	\caption{
		The proposed structure for skill transfer.
	} \label{Presentation2}
\end{figure}

This section introduces the proposed platform for the creation of the predicted-constructed path. This platform incorporates a deep convolutional neural network to generate a reference model while proficiently assessing the trainee surgeon's movements. Additionally, this structure employs two datasets, which are described in the following sections, while the second one is developed in this study. Starting with the trainee (novice) surgeon, skill characteristics are injected into the primary path automatically generated for him/her. This is accomplished through the reference model developed by the expert surgeon's feature space, in order to create an expert predicted-constructed path. \autoref{Presentation2} illustrates an overview of this process. 
\subsection{Datasets}
Two datasets are used in this study to consider both general and specific surgical skills. The first dataset, designated JIGSAWS, was created by Johns Hopkins University. The second dataset, dubbed as ARAS-Farabi (Cataract-LMM) \cite{cataract_lmm}, is developed in this paper and was created through a partnership between the Farabi Hospital and the ARAS research group. To augment datasets, various methods can be utilized, such as using different time frames of surgical data and employing different permutations of Cartesian coordinates. For instance, interchanging the data related to x and y coordinates ([x,y] to [y,x]) does not affect the comprehension of the skill represented by the data sequence. In the next sections, we will provide a brief overview of the two datasets.

\subsubsection{JIGSAWS Dataset}
\begin{figure}[h!]
	\centering
	\subfigure[Needle-Passing]{
		\includegraphics[width=0.3\columnwidth]{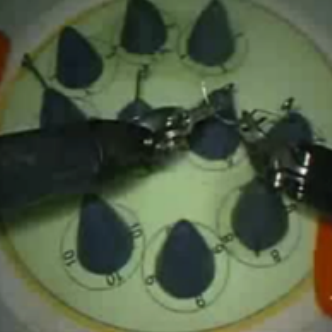}
		\label{fig:3FDMs3a2vin2gsCW1}
	}
	\subfigure[Suturing]
	{
		\includegraphics[width=0.3\columnwidth]{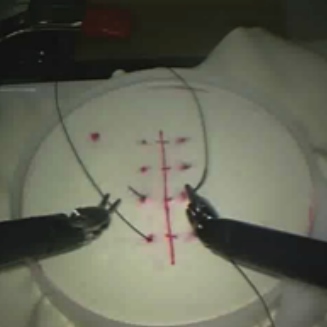}
		\label{fig:3FDMsavi2ngC2sW2}
	}
	\subfigure[Knot-Tying]{
		\includegraphics[width=0.3\columnwidth]{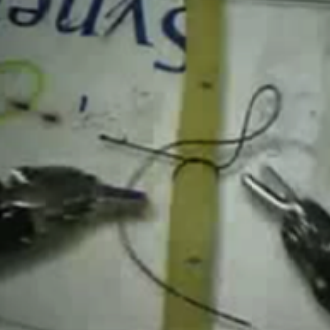}
		\label{fig:3FDMs3a2vin22gsCW1}
	}
	\caption{
 Three general surgical tasks in JIGSAWS dataset.
	} \label{dav5inci}
\end{figure}

JIGSAWS stands for JHU-ISI Gesture and Skill Assessment Working Set.
This dataset has collected 76 kinematic variables from the movements of 103 surgeons.
These data are obtained from three surgical tasks of suturing, needle passing, and knot tying, while the skill level of the surgeons who performed these tasks is also included in this dataset~\cite{c9,c10}. 
These data were taken from the da Vinci robotic surgery system, and the skill annotations mostly depend on hours of surgical experience. An image related to the JIGSAWS dataset is shown in \autoref{dav5inci}. In this paper we use the kinematic variables $x, y,$ and $z$ (position) from this dataset.

\subsubsection{ARAS-Farabi Software and Dataset}
There are several challenges in teaching and learning cataract surgery while preserving patient safety. One of the limitations of intraocular surgery teaching is that the attending surgeon mostly plays the role of an observer/supervisor during the surgery and can’t prevent potential complications. Moreover, surgical performance needs deep knowledge of ocular anatomy, manual dexterity, and mastering basic surgical skills \cite{c102221}. A continuous curvilinear anterior capsulotomy (CCC, capsulorhexis) is one of the critical initial steps in this surgery. The goal of capsulorhexis is to create a circular opening in the anterior surface of the cataracted lens. Creating this opening is one of the most challenging tasks for first-year residents \cite{c102222, c102223, c102224}. Therefore, creating a dataset of capsulorhexis surgeries is important.

On the other hand, as shown in \autoref{dav5inci}, the JIGSAWS dataset is far from representing a real surgery process, and research that solely relies on it will not significantly contribute to the development of effective and realistic surgical training. To address this issue, this study has developed a dataset of actual surgical procedures, which is introduced in this section.
In cataract surgery, interventions are performed on the human eye using tiny tools under a surgical microscope. The surgical microscope camera captures the video that the surgeon sees through the eyepiece. The microscope camera records surgical videos with pixel dimensions of $720 \times 576$ (using the H.264/AVC codec). The surgical videos were recorded and then pre-processed to enable annotation and extraction of motion data. The data sampling rate, which falls within the range of 25 to 30 data points per second, has been carefully selected to capture sufficient information to acquire skill features. The amount of input data has been also calculated with consideration of providing ample time for the acquisition of such features.

ARAS-Farabi dataset contains more than one hundred capsulorhexis surgery recordings. Initially, three expert surgeons annotated the recordings based on distinct skill indicators and classified them as expert or novice. Subsequently, the motion data related to surgical videos were retrieved using the ARAS-Farabi surgical analysis software~\cite{c7}.
This software is developed with the Python programming language and the PyQt library, and it is available in two versions for Windows and Linux. Although ARAS Group is trying to improve this software's tracking tools \cite{cite1}, it currently relies on OpenCV library trackers. In this software, any tracking algorithms from the OpenCV library can be utilized; however, the best method based on an objective evaluation of the ARAS-Farabi dataset is specified as the default mode. 

This program is user-friendly and can be used for various medical purposes. It is sufficient for the user to open the video and identify the main surgical areas. 
This software allows human intervention as an observer. It is possible to redefine the bounding boxes by pausing the video and going back if the tracker performance needs to be corrected. This software calculate position, speed, and acceleration of the centers of the two main areas of the surgery (absolutely and relatively), stored, and displayed in real-time as graphic diagrams and numerical data.
\begin{figure}[h!] 
	\centering
	\includegraphics[width=\columnwidth]{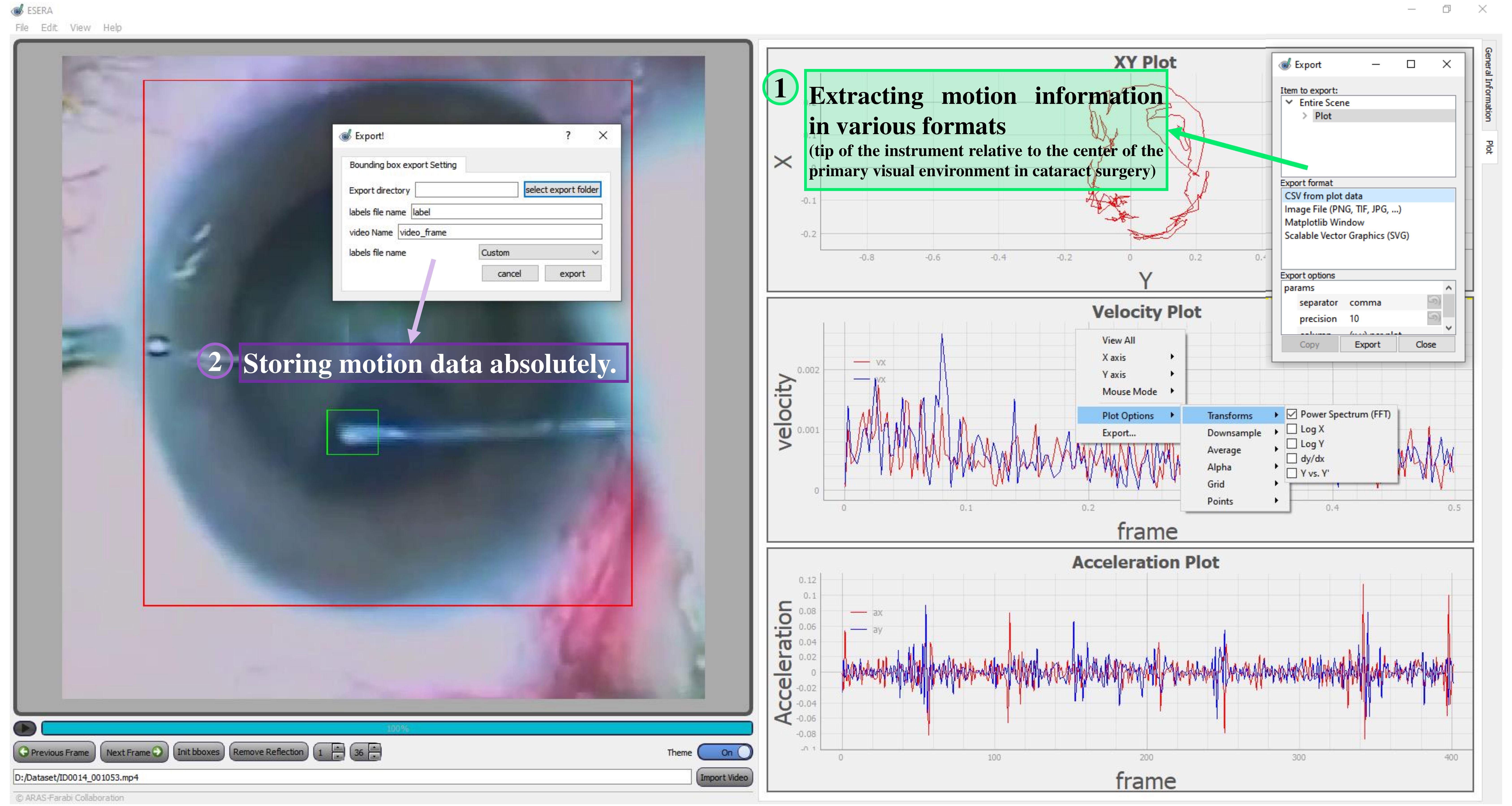}
	\caption{Output information of Aras-Farabi software.(\href{https://drive.google.com/file/d/12bP5Y8SWm3enYijG5Gh5QCj0rSNzqt_k/view?usp=sharing}{Video}).
	} \label{APP110}
\end{figure}
This software addresses some challenges associated with extracting motion data from surgical videos. 
For example, intraocular light reflection may mislead computer vision algorithms into tracking the tip of a capsulorhexis instrument that is visually close to this reflection. 
As shown in \autoref{APP2}, this software has tracked and removed the light reflection in all video frames using morphology operations and OpenCV libraries.
To address the challenge of frame scales and the non-uniformity of video magnifications, all motion information is scaled according to the length of the pupil's bounding box, which is nearly the same for all individuals according to our goals \cite{cite2}.
\begin{figure}[h!] 
	\centering
	\includegraphics[width=\columnwidth]{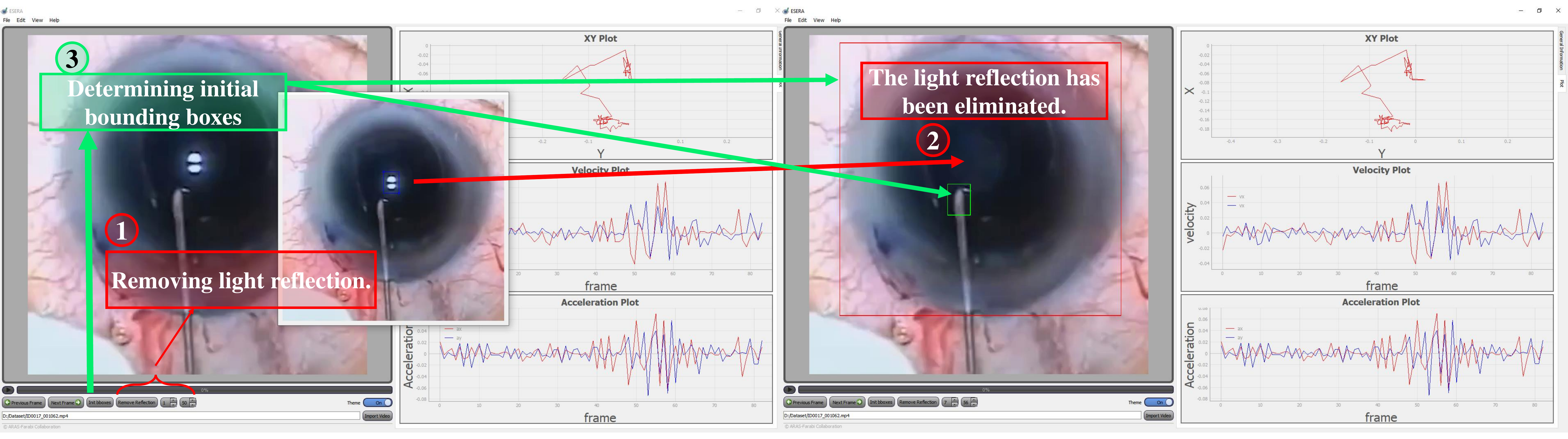}
	\caption{Removing light reflections and defining initial bounding boxes (\href{https://drive.google.com/file/d/1NaURHEClxTbTA2-\_chnLNDwy2wMUJX-G/view?usp=sharing}{Video}).	
	} \label{APP2}
\end{figure}

\begin{figure}[h!] 
	\centering
	\subfigure[{X1}]{
		\includegraphics[width=0.47\columnwidth]{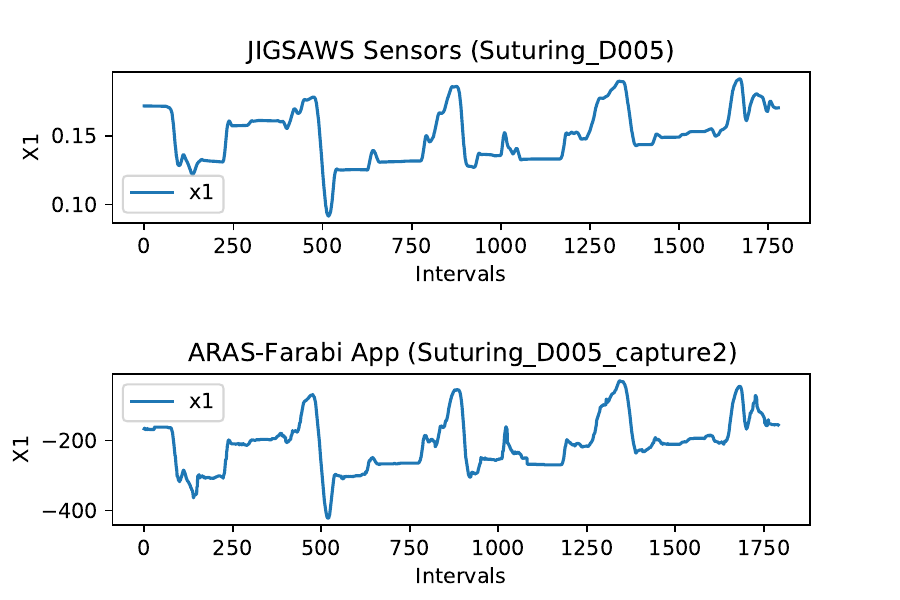}
		\label{fig:FDMs1avin3g23240000C0W1}
	}
	\subfigure[{Y1}]
	{
		\includegraphics[width=0.47\columnwidth]{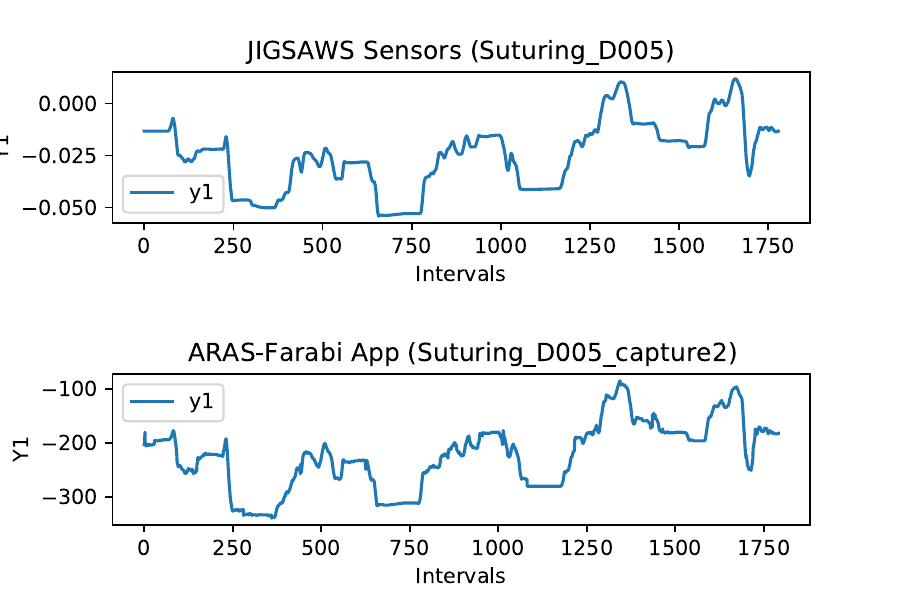}
		\label{fig:FDM1sav0i3n4223g0000CW2}
	}
	\caption{
 Comparison of information extracted from ARAS-Farabi software with sensory data.
	} \label{trac332388333ker1}
\end{figure}

\autoref{trac332388333ker1} compares the 2D position information for the two tools in a surgery recording of the JIGSAWS dataset to validate whether the motion information was retrieved using this tracker-based software is genuinely related to the accurate sensory information. The JIGSAWS dataset was chosen for this section because it includes sensing data synced with the video. The upper subgraphs of each data point are retrieved from the sensor, and the lower subgraphs of each graph are extracted using the ARAS-Farabi program. The ARAS-Farabi software can keep the surgeon's path going in its output data with high accuracy. It merely requires the appropriate scale to map to the sensor data. It should be noted that the general moving process rather than the precise, scaled numerical information is essential in assessing surgical skills. With this software, a data set of surgeon movement information in sync with the videos was created.

This study employs position and velocity data retrieved from videos of capsulorhexis surgery.
\subsection{Skill-related feature extraction}
Creating a knowledge repository of surgical skills is crucial for skill transfer. Developing an artificial intelligence model for differentiating and categorizing surgeons' skills into experts and novices enables the model to comprehend the features of expert surgeons' motions.
The model's trained and stored weights, parameters, and layers are then used to embed these features. These models are trained on JIGSAWS \cite{c9,c10} and ARAS-Farabi datasets to infer both general and specific surgical skills. With more manageability in the prediction and motion route generation phase, an AI model that uses less computational resources for feature extraction will be more desirable.

Our neural network model was developed based on research that suggests skilled surgical performance can be assessed by analyzing hand movements \cite{c101}.  Rather than using all of the motion information in Cartesian coordinates as input, only a portion was utilized. The remaining data patterns were extracted using neural network filters.
Due to these considerations, our network is light, prepared for quick processing, and able to decide based on a small subset of the data. In the following, the architecture of the proposed network is introduced. The input to the network consists of N data points (intervals) of absolute or relative x, y (and z or v) coordinates. The smoothness of the entry path is a key aspect of the surgeon's skill, according to our research on the evaluation of surgical performance \cite{c7}.  

\begin{figure}[h!] 
\centering
\includegraphics[width=\columnwidth]{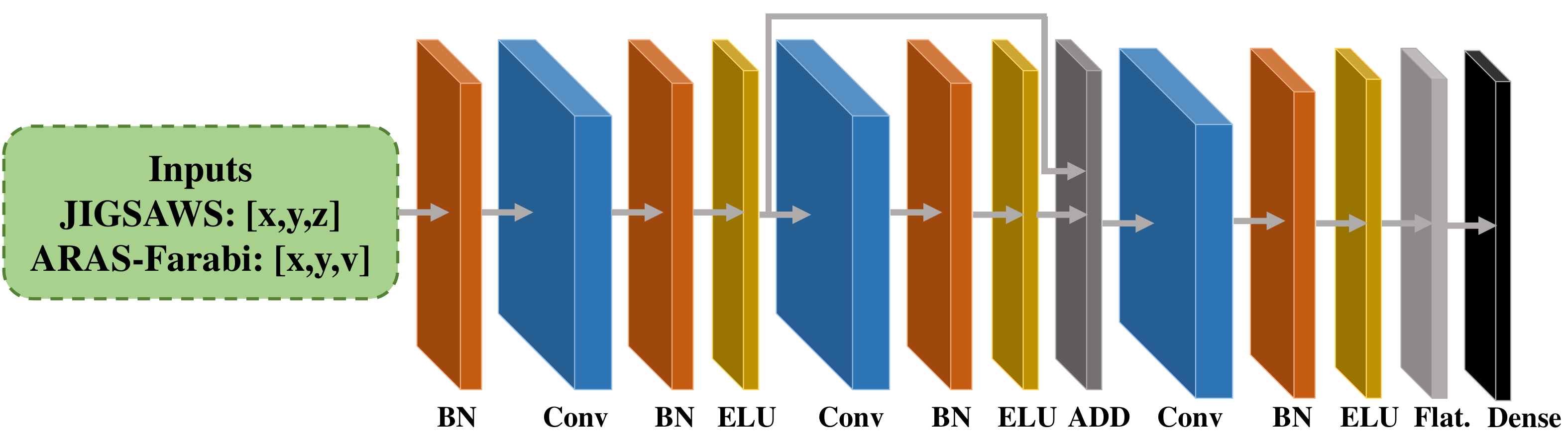}
	\caption{
		The architecture of the proposed network.
	} \label{ddPicture4}
\end{figure}

It has been shown in several studies that spatiotemporal information such as movement fluidity and path smoothness can be effectively obtained in the frequency domain \cite{c7, c770, c771}. 
Therefore, the input data points are passed through an Fast Fourier Transform (FFT) block to utilize the frequency domain. This is the input that the network receives.
The general structure of the network comprises convolutional layers, batch normalization layers, and ELU activation functions.
\autoref{ddPicture4} depicts an overview of this network. 
It is notable that the task of extracting significant features is left to be handled by the convolutional network itself, rather than adding more information to the input.
Patterns from the input data, including those that are not immediately apparent to us and those that are intelligible, such as speed and acceleration, are extracted by convolutional layers.
In \autoref{ddPicture4}, it can be observed that the output of the initial layers is connected to the subsequent layers, which allows the input information to be preserved throughout the depths of the network and utilized in the latent space.
It is postulated that if the weights, coefficients, and hyper-parameters of the structure are adjusted to the extent that it can differentiate between skilled and novice surgery in its final fully connected layer, then the pre-classification layer of the structure will possess a latent space of skilled and novice surgeons' movement patterns.

\subsection{Predictive-constructive  }
Utilizing optimization techniques and style transfer methods can enhance the proficiency of novice surgeons without compromising their surgical intent.
For this purpose, a function for bilateral optimization needs to be developed \cite{ccc4}.
A function is proposed to maintain the surgeon's intended path while appearing to have been taken with greater expertise. The relation for the loss function used in optimization is defined as: 
\begin{equation}\label{eq1}
{L}_{ {Total}}={a} {L}_{ {Decision}}(\vec{P}, \vec{N})+{b} {L}_{ {Skill }}(\vec{E}, \vec{P}).
\end{equation}
The amount of AI agent intervention required to construct and suggest an improved path from the original path can be determined based on the distribution of coefficients $a$ and $b$, and the duration of the novice surgeon's training.
It can be conceived that at the onset of training, when a novice surgeon has less expertise, the ratio of $\frac{a}{b}$ may be greater than one, and that over time, this ratio can be decreased to a value between zero and one.

The novice surgeon's entry path is aimed to be improved with minimal deviation from their intention through the utilization of the decision loss function.
This loss function's minimization corresponds to minimizing the Euclidean distance between the points on the novice surgeon's path and the suggested path. By utilizing the concept of feature vectors instead of point distance between the paths, attention can be directed toward the higher-level features of movements.

An appropriate loss function for the skill needs to be defined to improve the movement path of the novice surgeon.
This function is capable of operating such that the statistical distribution of the feature space of the input path is comparable to that of the hidden layers of expert surgeons. 
To achieve this, the maximum mean square difference (MMD2) can be utilized to minimize the differences between statistical indicators, such as means, variances, and higher order moments, in the feature vector distribution of the input path and the expert path. In this study, the Gram matrices or the inner product between the distributions of novice and expert feature vectors and MMD2 are minimized.

The relationships of the decision and skill loss functions are defined in the form of equation \autoref{eq2}, in accordance with the justifications given.
\begin{equation}\label{eq2}
\begin{aligned}
& {L}_{{Skill }}(\vec{E}, \vec{P})=\left\|{Gram}_{f_E}-{Gram}_{f_P}\right\|_2 \\
& {L}_{ {Decission }}(\vec{P}, \vec{N})=\left\|\boldsymbol{f}_P-\boldsymbol{f}_{\boldsymbol{N}}\right\|_2
\end{aligned}
\end{equation}

\section{Results and Discussion}\label{s2}
The network is trained using the provided data in the first step, and a 99\% F1 score is obtained after tuning the parameters and hyperparameters. The surgical skill knowledge that is embedded in the network can be utilized by saving the model as an h5 file.
By using the proposed platform and considering the online path of the trainee (novice) surgeon during surgery, the suggested path is created. The suggested path is initialized with the novice path and constructed using the Adam optimization method with a learning rate of $\gamma$ as given in \autoref{eqq}.
\begin{equation}\label{eqq}
\vec{P}:=\vec{P}-\gamma \frac{\partial \mathcal{L}_{\text {Total }}}{\partial \vec{P}}
\end{equation}
The improvement in the skill level is then determined by passing both paths through the proposed network and comparing the skill percentage of each.
The skillfulness of the proposed path may not be quantifiable from an image or network output. To address this issue, quantitative criteria are employed to assess the output path. Experienced surgeons exhibit more fluid and predictable movements. A hybrid metric is employed, which considers the surgeon's back-and-forth hand motion and provides an estimate of motion entropy. The criterion value decreases as the surgeon's movements become more skilled and predictable. Furthermore, a combined measure of speed (energy consumption) and the Euclidean smoothness of frequency coefficients is also employed to evaluate motion smoothness.

\subsection{ARAS-Farabi Dataset}
\begin{figure}[!t]
	\centering
	\subfigure[$\frac{b}{a}=1.5$]{
		\includegraphics[width=0.45\columnwidth]{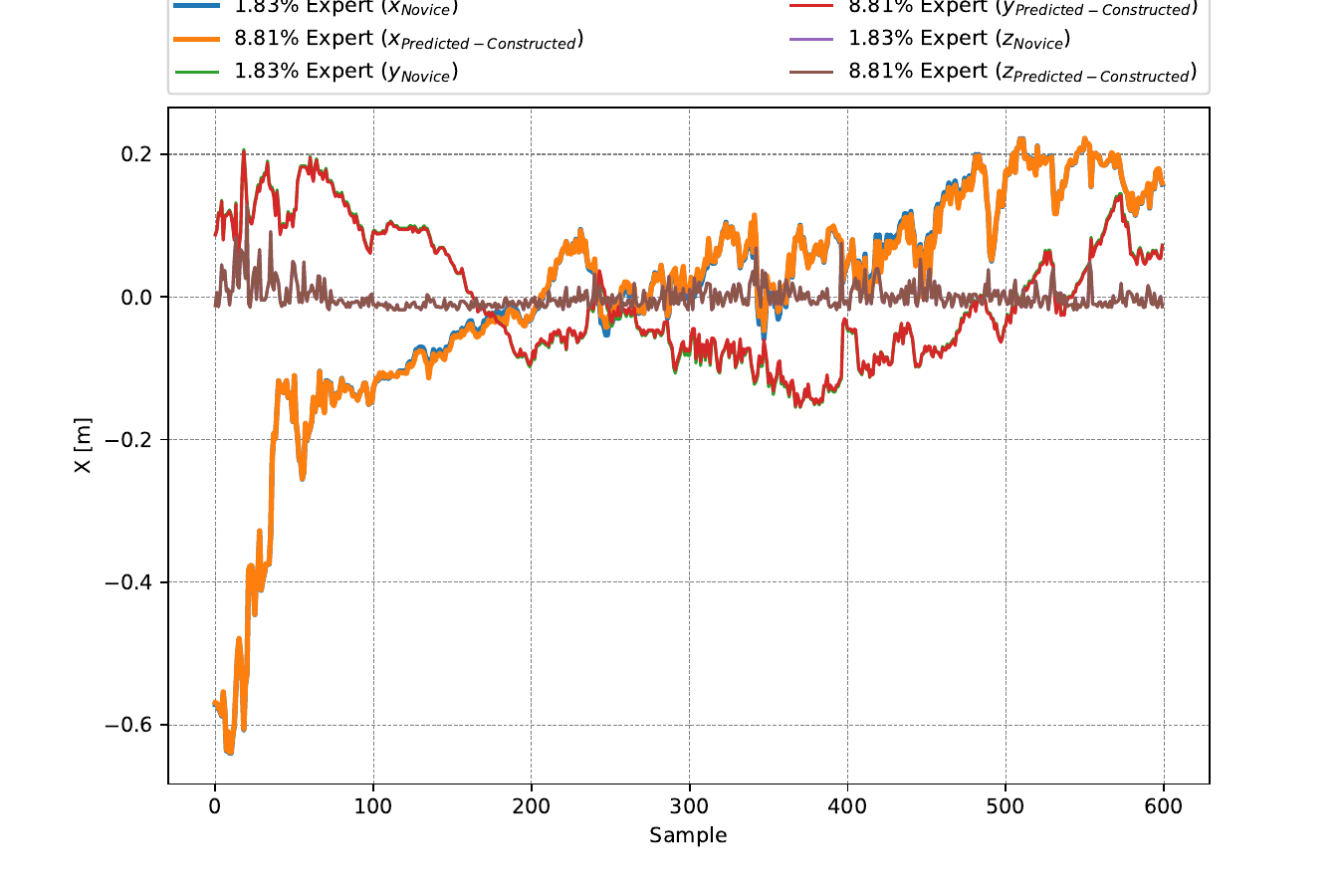}
		\label{fig55}
	}
	\subfigure[$\frac{b}{a}=1.5$]
	{
		\includegraphics[width=0.45\columnwidth]{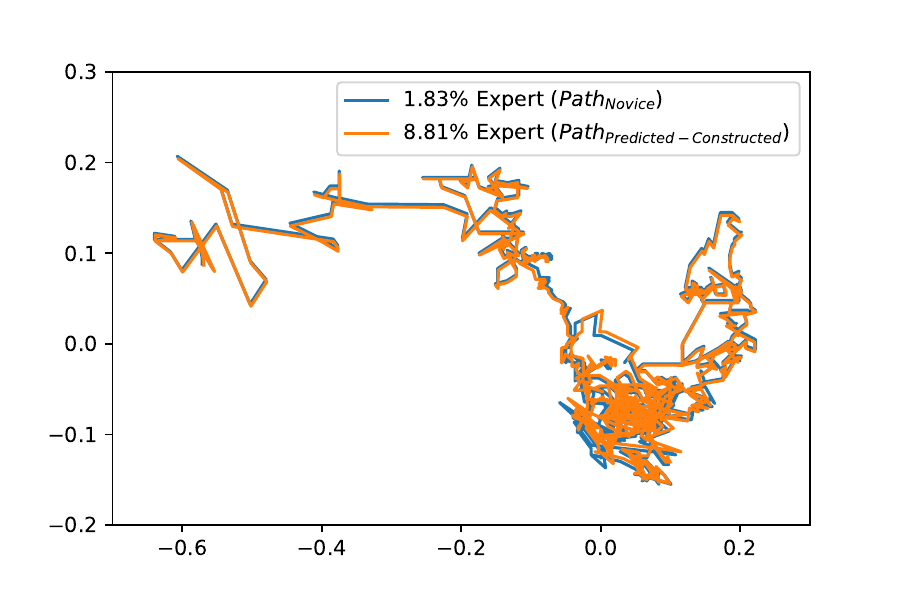}
		\label{fig56}
	}
 \\
 \subfigure[$\frac{b}{a}=5$]{
		\includegraphics[width=0.45\columnwidth]{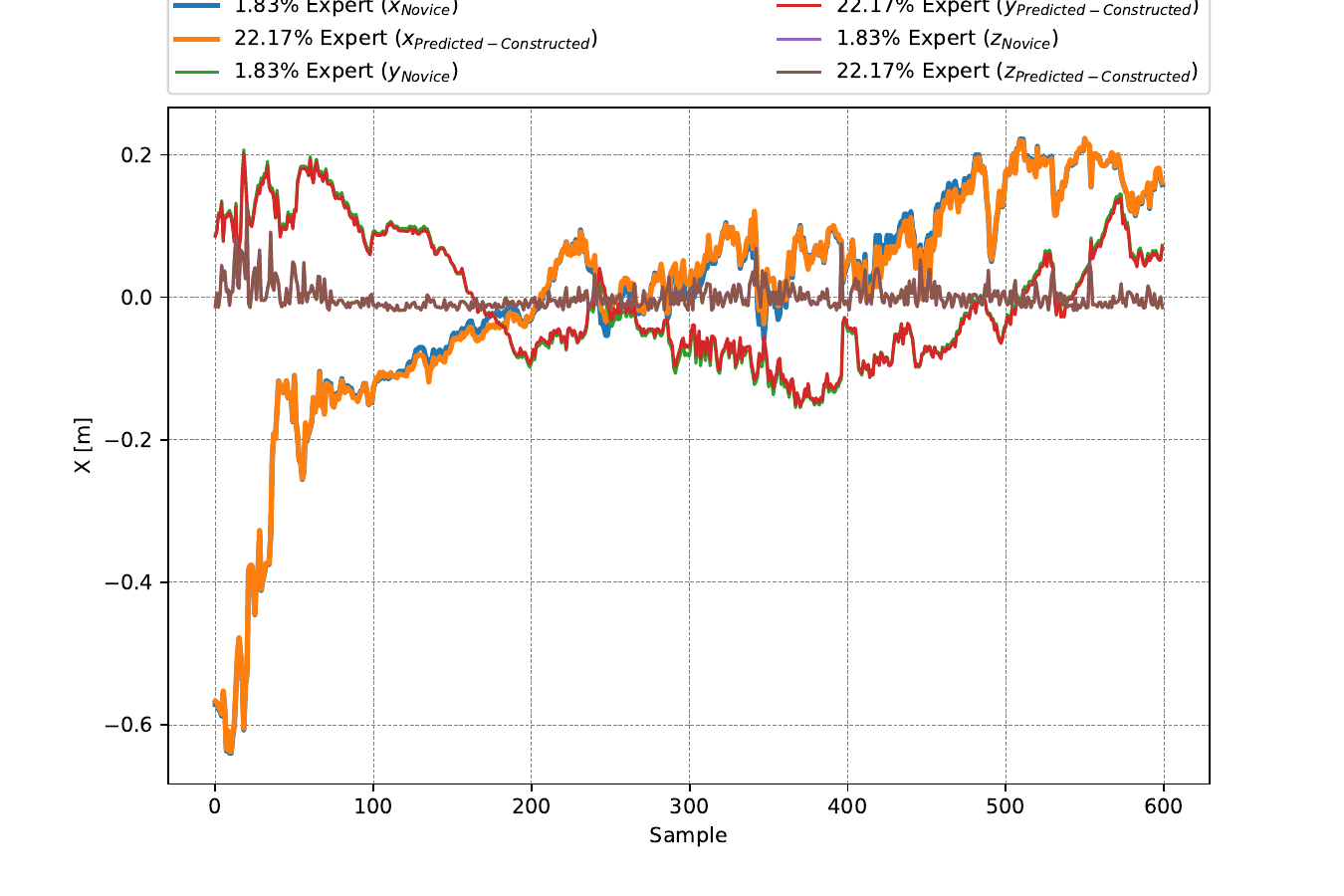}
		\label{fig525}
	}
	\subfigure[$\frac{b}{a}=5$]
	{
		\includegraphics[width=0.45\columnwidth]{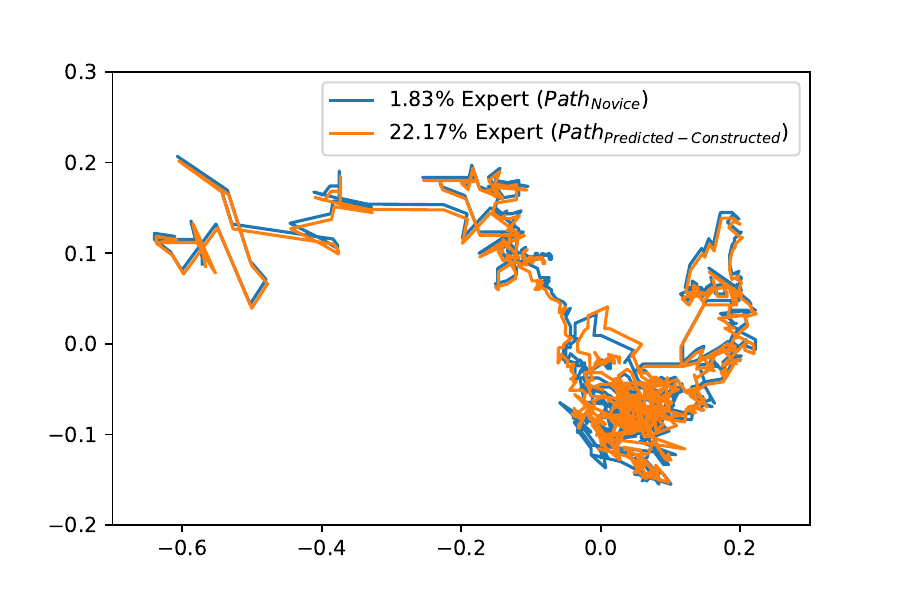}
		\label{fig526}
	}
	\caption{
		Comparison of the base path (novice) and the predicted-constructed path (expert).}
	\label{P122ddd22}
\end{figure}

The ARAS-Farabi dataset was examined to analyze the structure of skill transfer, with the use of scaled-relative position and speed data retrieved from the capsulorhexis cystotome tool-tip. To conduct the first experiment, an input path was selected based on the novice data from the ARAS-Farabi dataset. The features of an expert surgery were injected into the initial path from the surgical path of one of the surgeries labeled as the expert in the ARAS-Farabi dataset.
By varying the values of coefficients $a$ and $b$ in \autoref{eq1}, it is possible to control the relative importance of improving the novice surgeon's path and preserving their intended movements. 
As shown in \autoref{P122ddd22}, when the significance coefficient of the loss function associated with the skill increases, the suggested movement path becomes more expert.
Additional information about the amount of change in the skill level of the novice surgeon's path when using the proposed path is provided in \autoref{tab2}.
The proposed structure results in a novice surgeon's new path that contains more expert characteristics and fewer unexpected movements and tremors.
\begin{table}[!b]
\caption{Skill enhancement of the novice surgeon using the predicted-constructed path.}
    \label{tab2}
    \centering
    \begin{tabular}{|l|c|c|}
\hline Indicators and Ratios of Coefficients & $\frac{\boldsymbol{b}}{\boldsymbol{a}}=\mathbf{1 . 5}$ & $\frac{\boldsymbol{b}}{\boldsymbol{a}}=\mathbf{5}$ \\
\hline Total Skill Level Improvement & $6.97$ & $20.33$ \\
\hline Total Predictability Improvement & $14.34$ & $18.87$ \\
\hline Enhancement in Tremor Reduction & $0.23$ & $3.44$ \\
\hline Enhancement in Noise Cancellation & $0.23$ & $2.40$ \\
\hline
\end{tabular}
\end{table}

\begin{figure}[!t] 
 	\centering
 	\includegraphics[width=0.9\columnwidth]{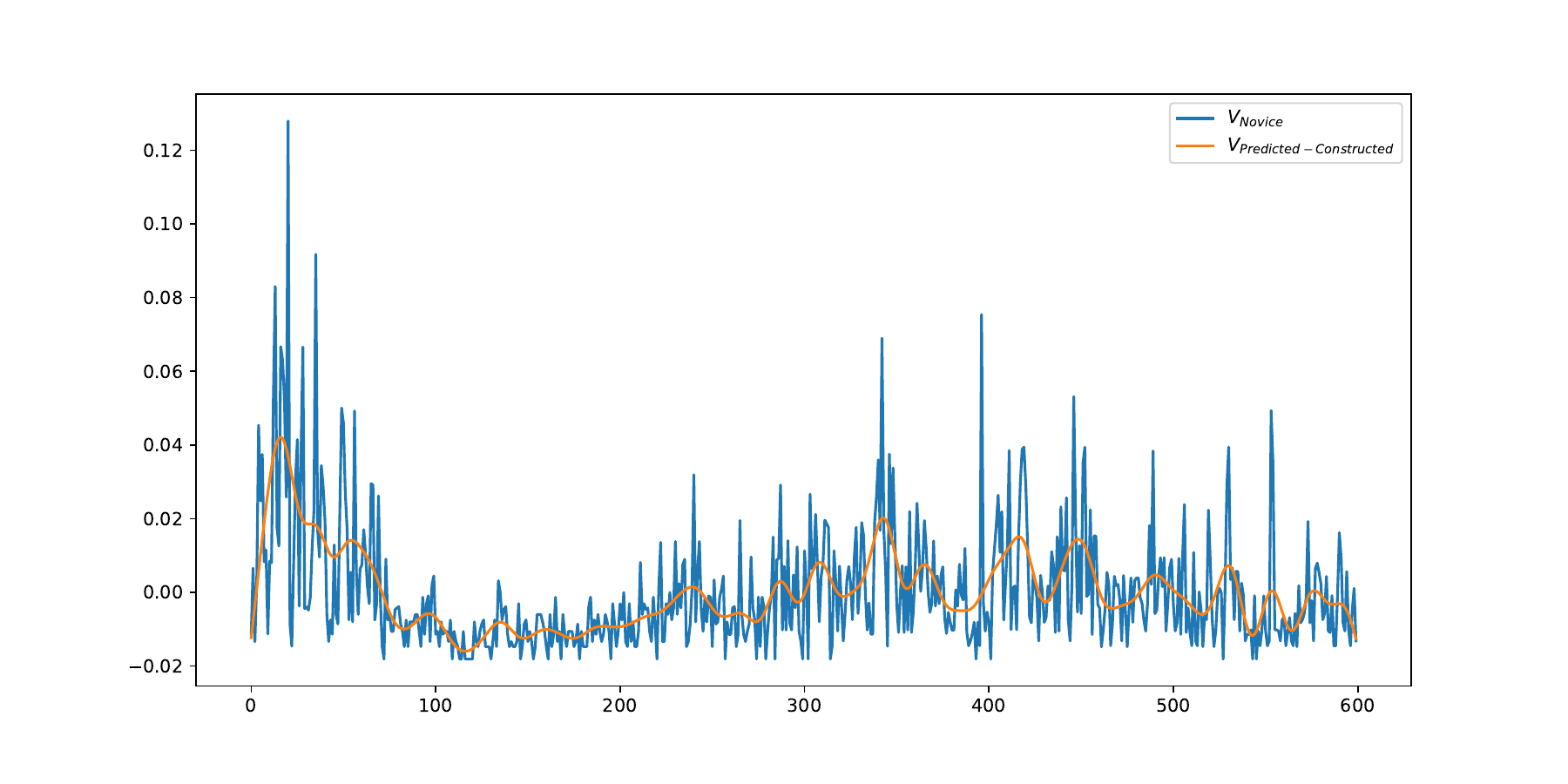}
 	\caption{Idealized performance of the proposed structure on the velocity dimension.
 	} \label{resultV}
 \end{figure}

The idealized performance of the proposed structure on the velocity dimension is depicted in \autoref{resultV}. It can be observed that the surgical instrument moves more smoothly with this architecture, thereby reducing risks and increasing the surgeon's skill level.
 
\begin{figure}[!b]
	\centering
	\subfigure[$\frac{b}{a}=1.5$]{
		\includegraphics[width=0.45\columnwidth]{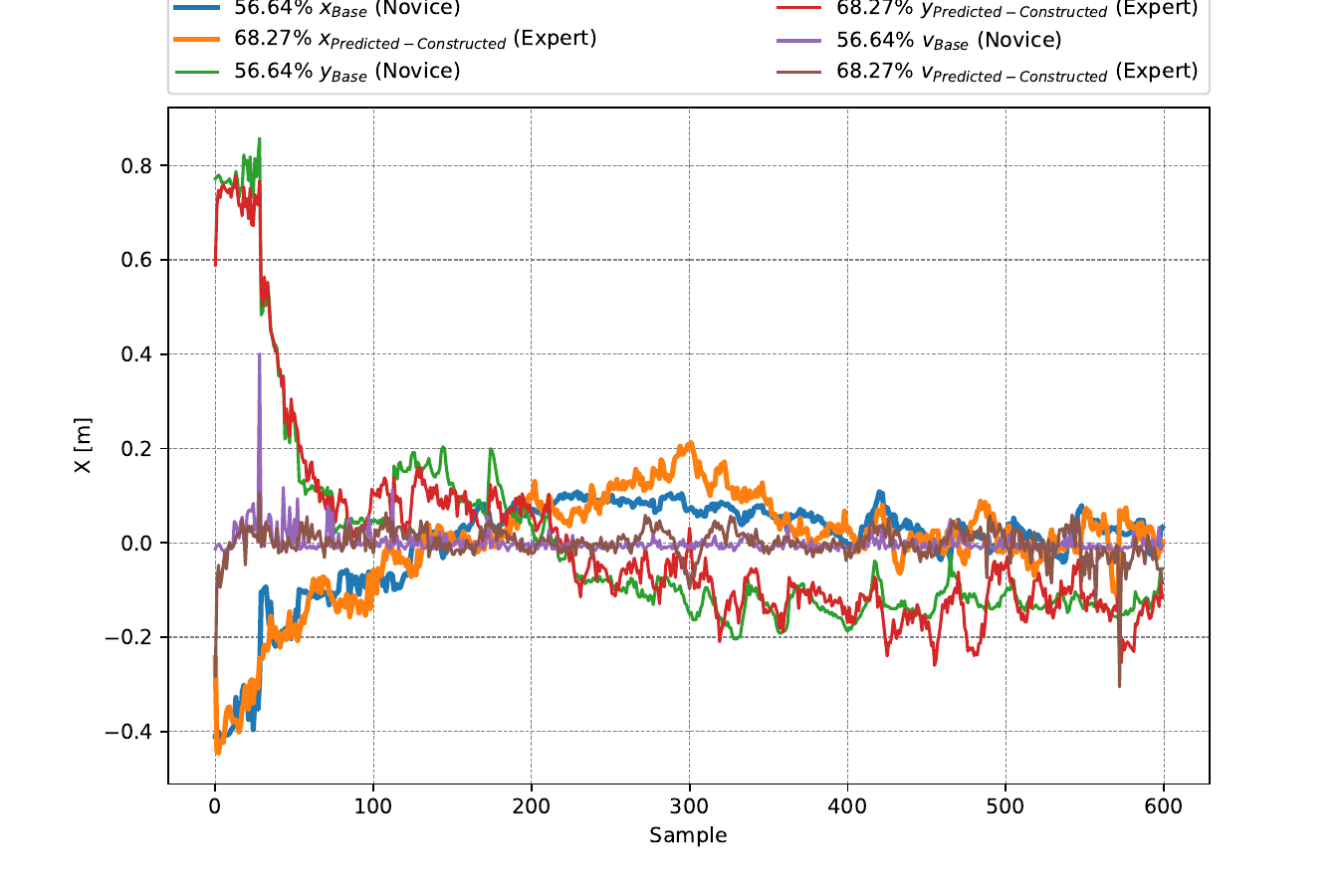}
		\label{fig1}
	}
	\subfigure[$\frac{b}{a}=1.5$]
	{
		\includegraphics[width=0.45\columnwidth]{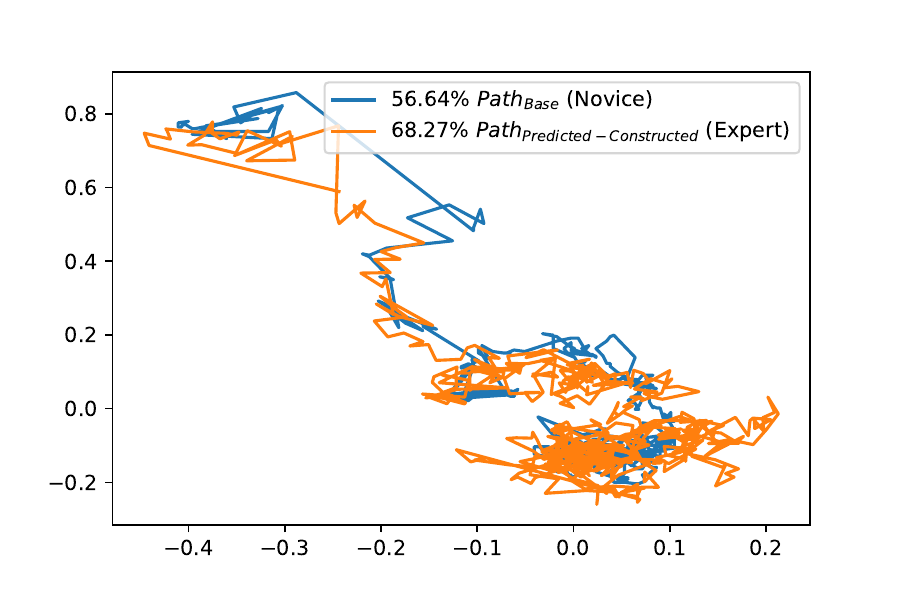}
		\label{fig2}
	}
 \\
 \subfigure[$\frac{b}{a}=100$]{
		\includegraphics[width=0.45\columnwidth]{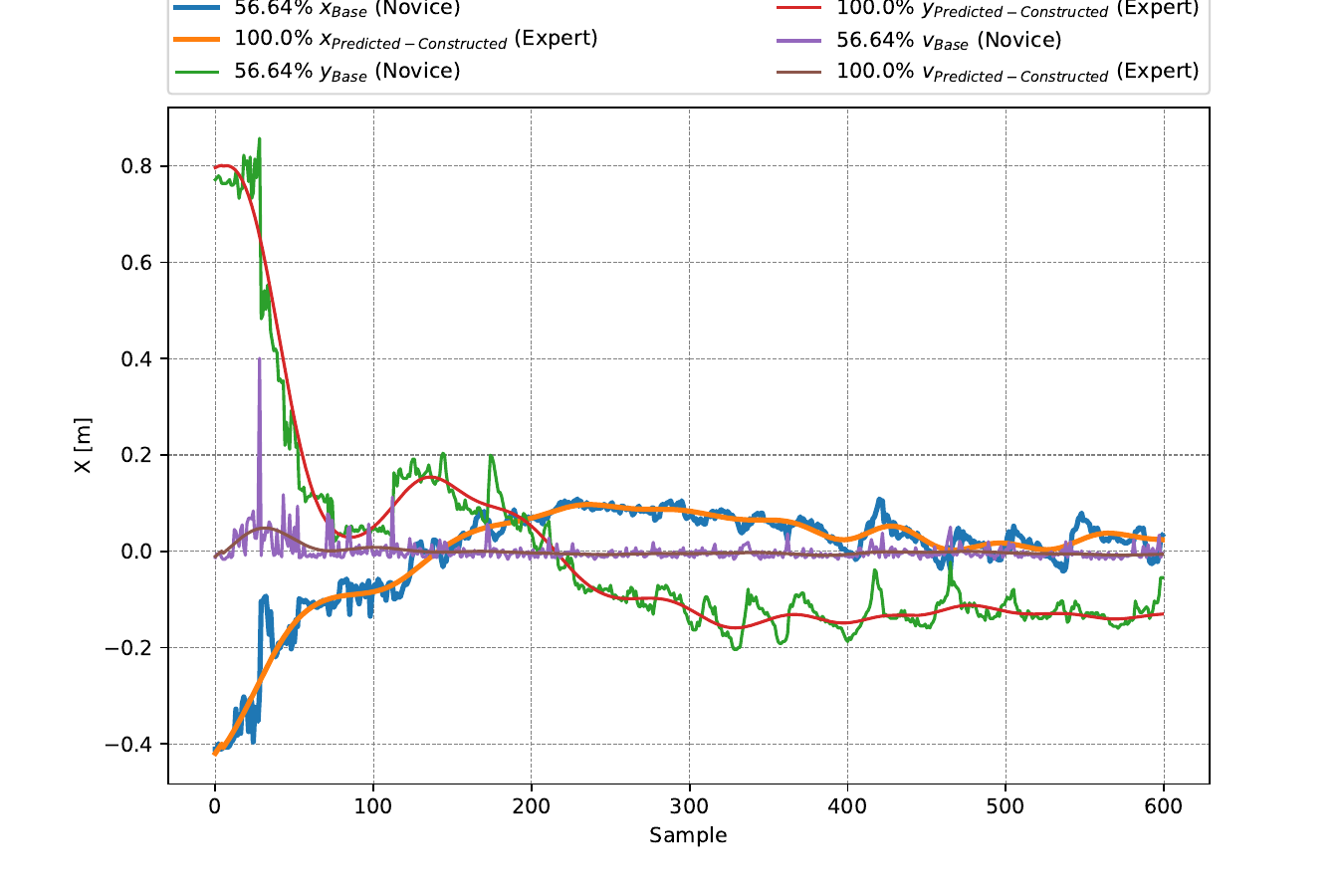}
		\label{fig3}
	}
	\subfigure[$\frac{b}{a}=100$]
	{
		\includegraphics[width=0.45\columnwidth]{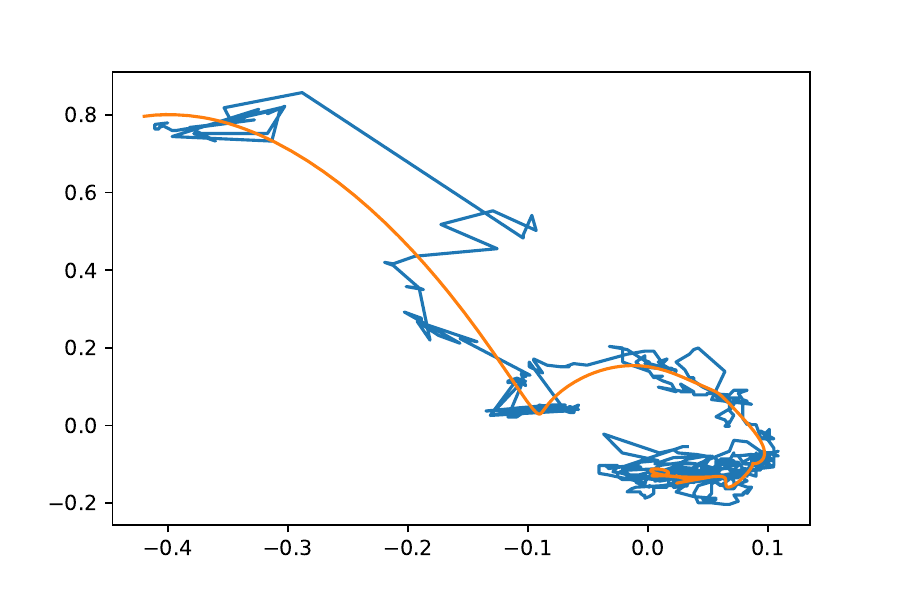}
		\label{fig4}
	}
	\caption{
		Comparison of the base path (novice) and the predicted-constructed path (expert).}
	\label{Pss1}
\end{figure}

 The proposed structure's performance is evaluated for a new path taken by a novice. When these coefficients are equally considered, it is observed that the skill level of the movement path is increased by approximately 11\%, while preserving the novice surgeon's general intention (\autoref{fig1} and \autoref{fig2}).
The rate of skilled improvement of the movement path will be about 41\% as we increase the important coefficient of the skill loss function (\autoref{fig3} and \autoref{fig4}). As can be observed in \autoref{Pss1}, the surgeon's actions (including positions and velocities) are now more fluid and less abrupt, in addition to having movement characteristics that have more expertise in terms of the trained network.

\begin{figure}[!t]
	\centering
	\subfigure[$\frac{b}{a}=5$]{
		\includegraphics[width=0.45\columnwidth]{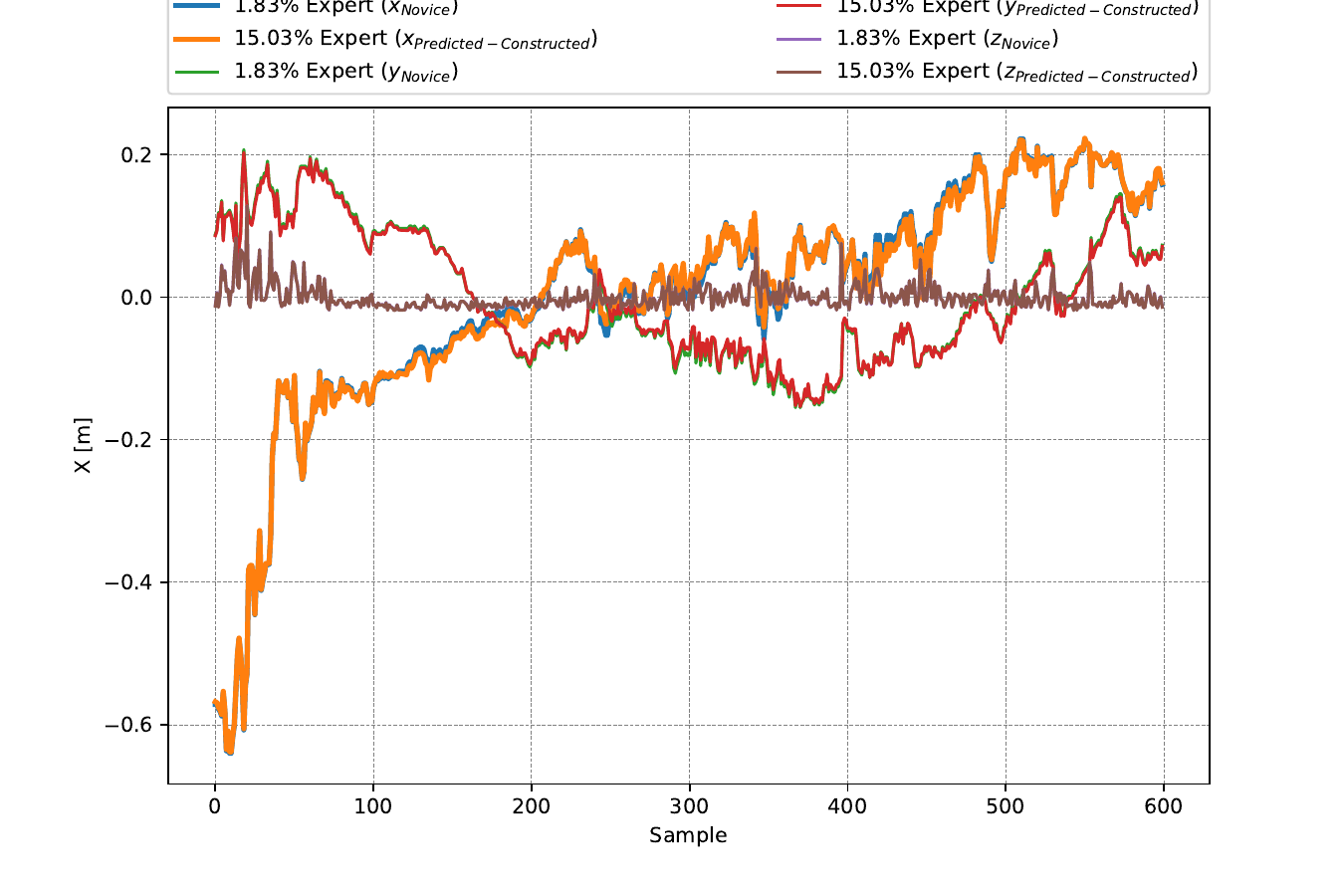}
		\label{fig5}
	}
	\subfigure[$\frac{b}{a}=5$]
	{
		\includegraphics[width=0.45\columnwidth]{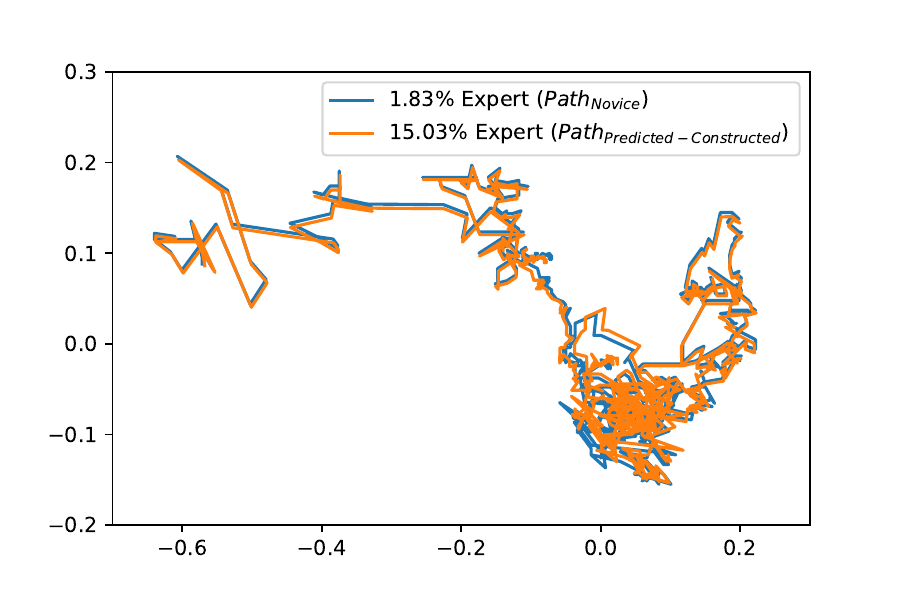}
		\label{fig6}
	}
	\caption{
		Comparison of the base path (novice) and the predicted-constructed path (expert).}
	\label{Psss12}
\end{figure}

The next experiment investigates the impact of employing various sets of features. The trainee's path is improved even further when the network injects skill behaviors extracted from a more experienced trainer. In other words, students will benefit from better instruction from a more experienced instructor.
To demonstrate this point, the characteristics of a less expert surgical path are used, and the proposed path under identical conditions improves the skill level by 7\% less. Both \autoref{Psss12} and \autoref{tab33} illustrate this fact.
\begin{figure}[!t]
	\centering
	\subfigure[ ]{
		\includegraphics[width=0.45\columnwidth]{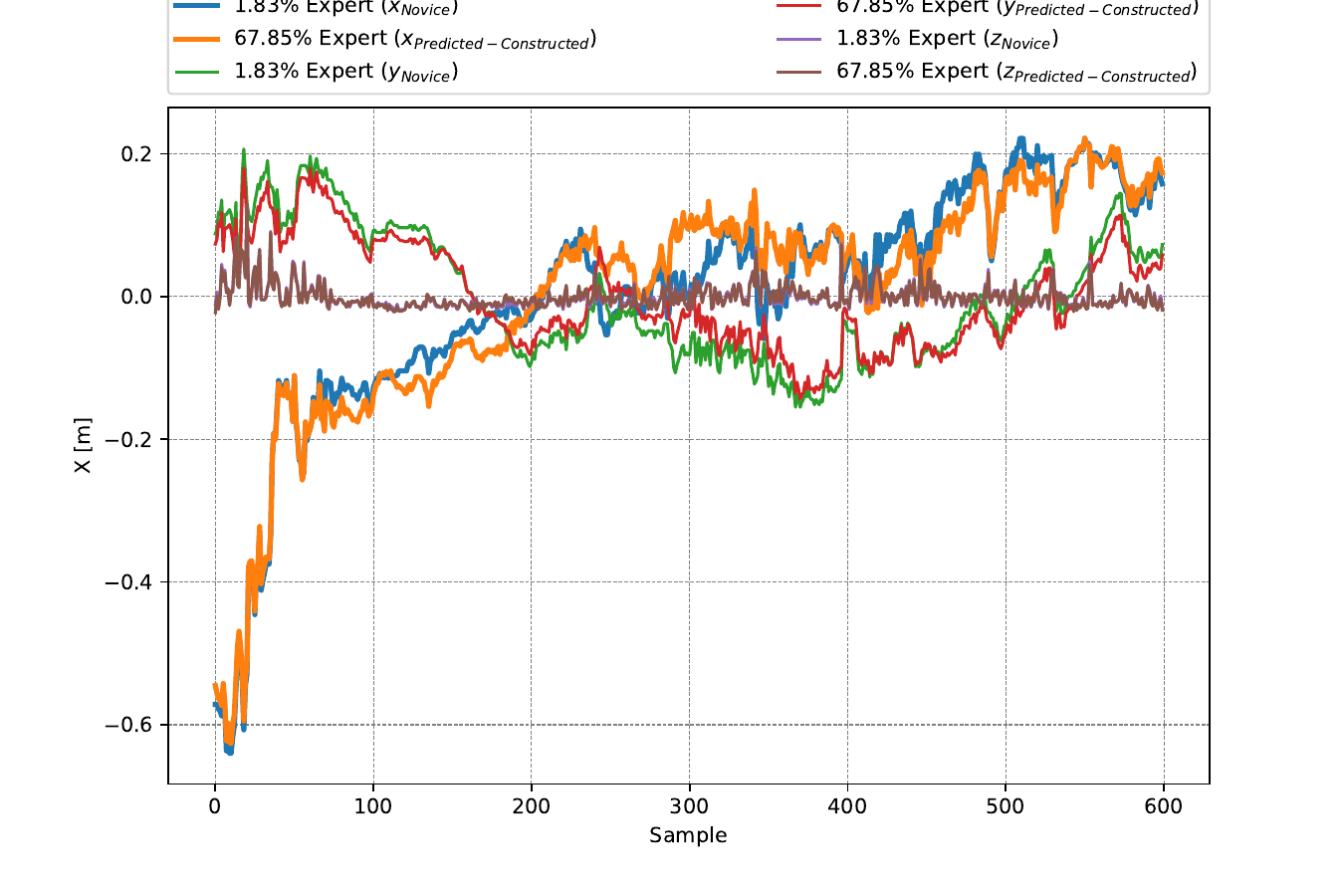}
		\label{fig5}
	}
	\subfigure[ ]
	{
		\includegraphics[width=0.45\columnwidth]{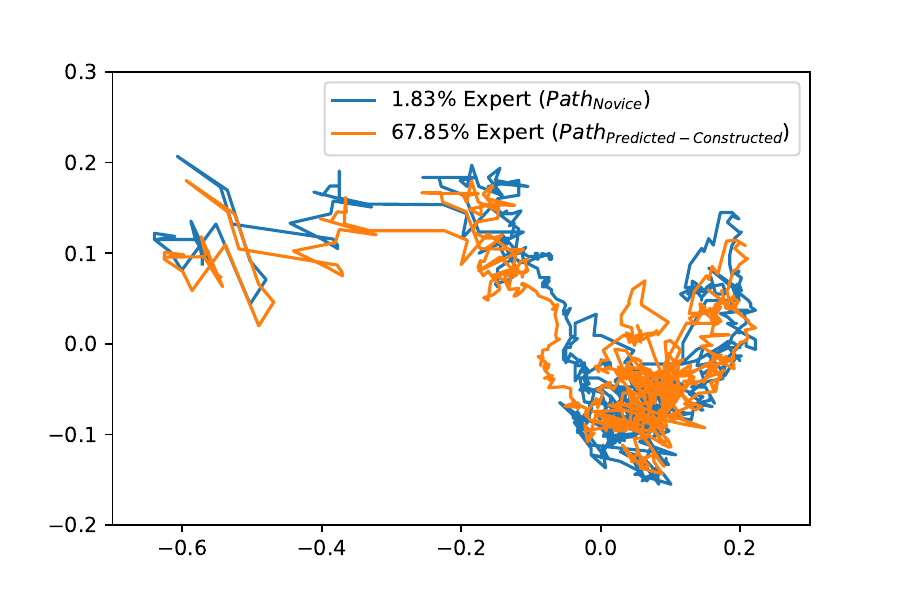}
		\label{fig6}
	}
 \\
 	\subfigure[ ]{
		\includegraphics[width=0.45\columnwidth]{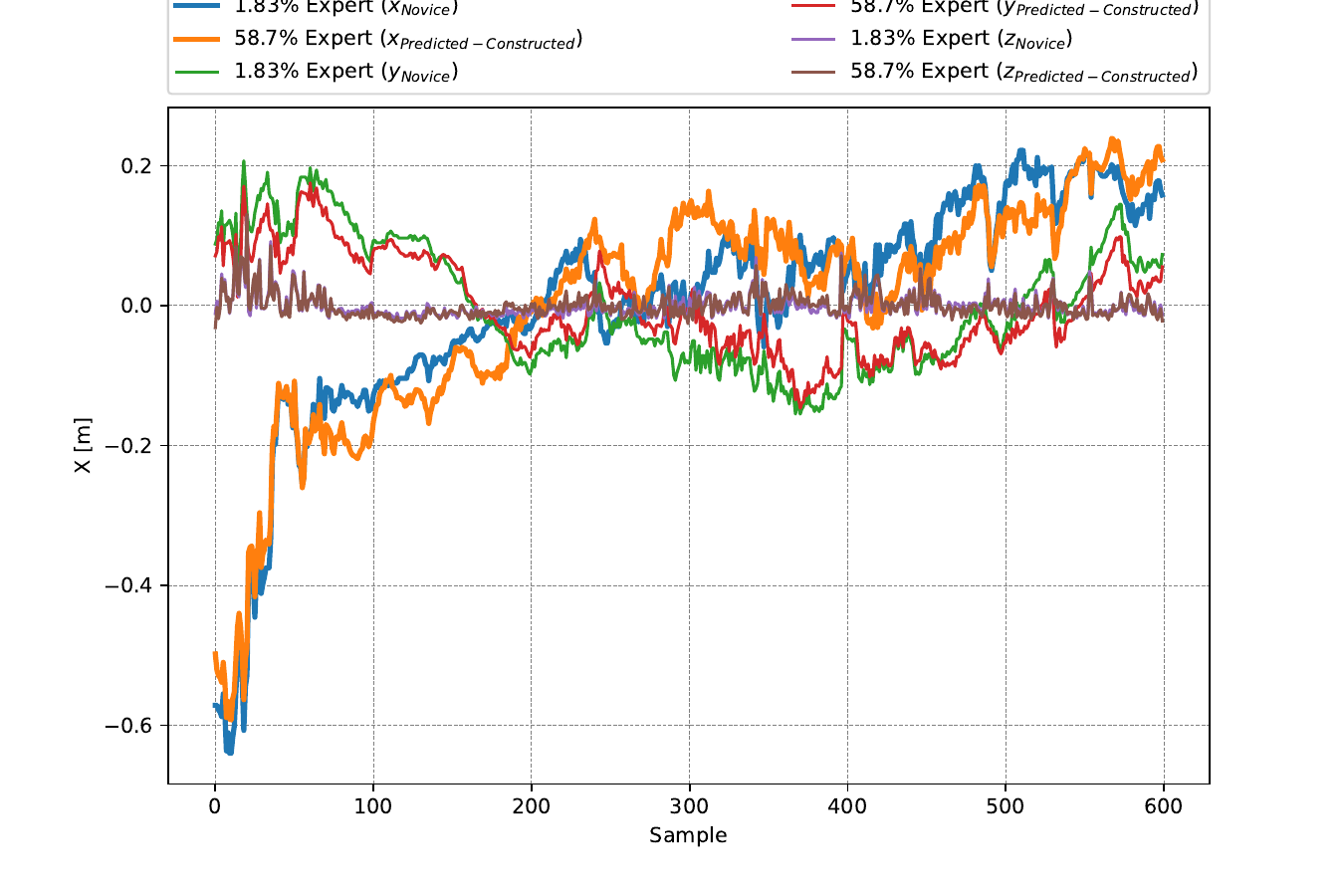}
		\label{fig5}
	}
	\subfigure[ ]
	{
		\includegraphics[width=0.45\columnwidth]{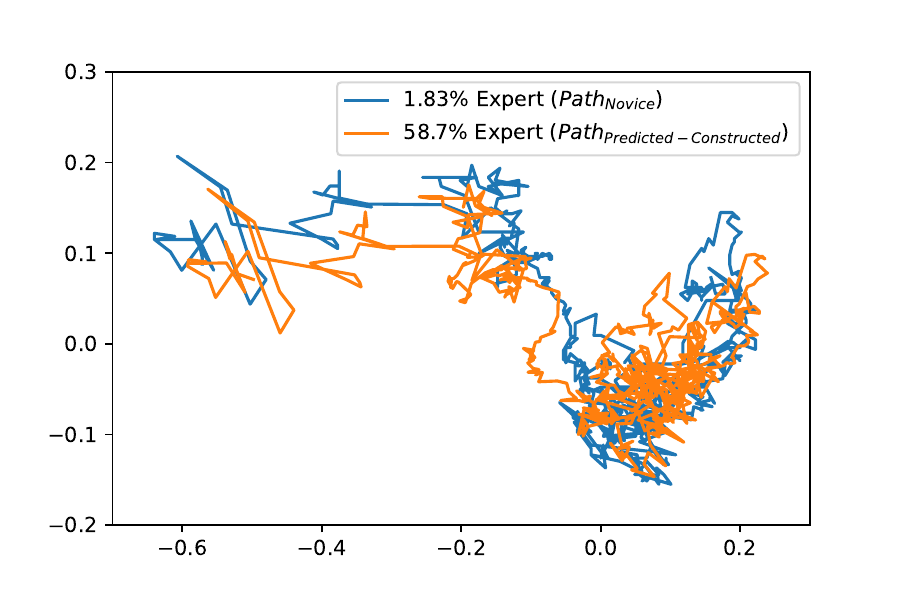}
		\label{fig6}
	}
 \\
 	\subfigure[ ]{
		\includegraphics[width=0.45\columnwidth]{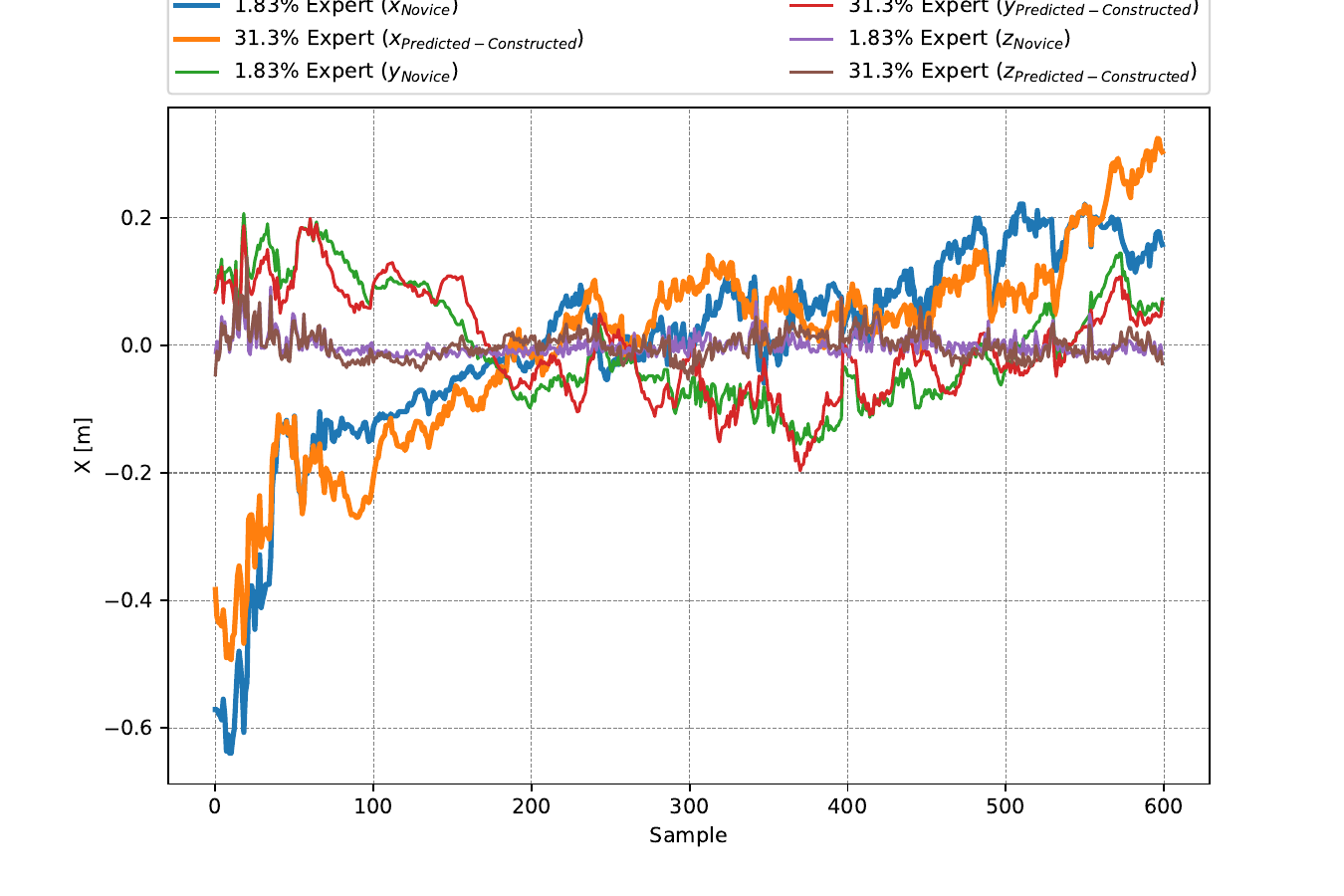}
		\label{fig5}
	}
	\subfigure[ ]
	{
		\includegraphics[width=0.45\columnwidth]{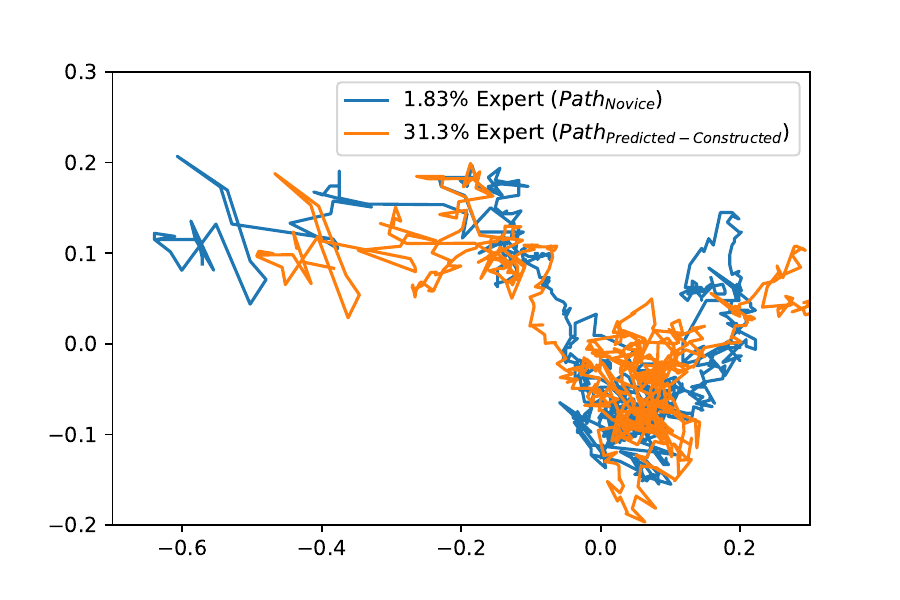}
		\label{fig6}
	}
 \\
 	\subfigure[ ]{
		\includegraphics[width=0.45\columnwidth]{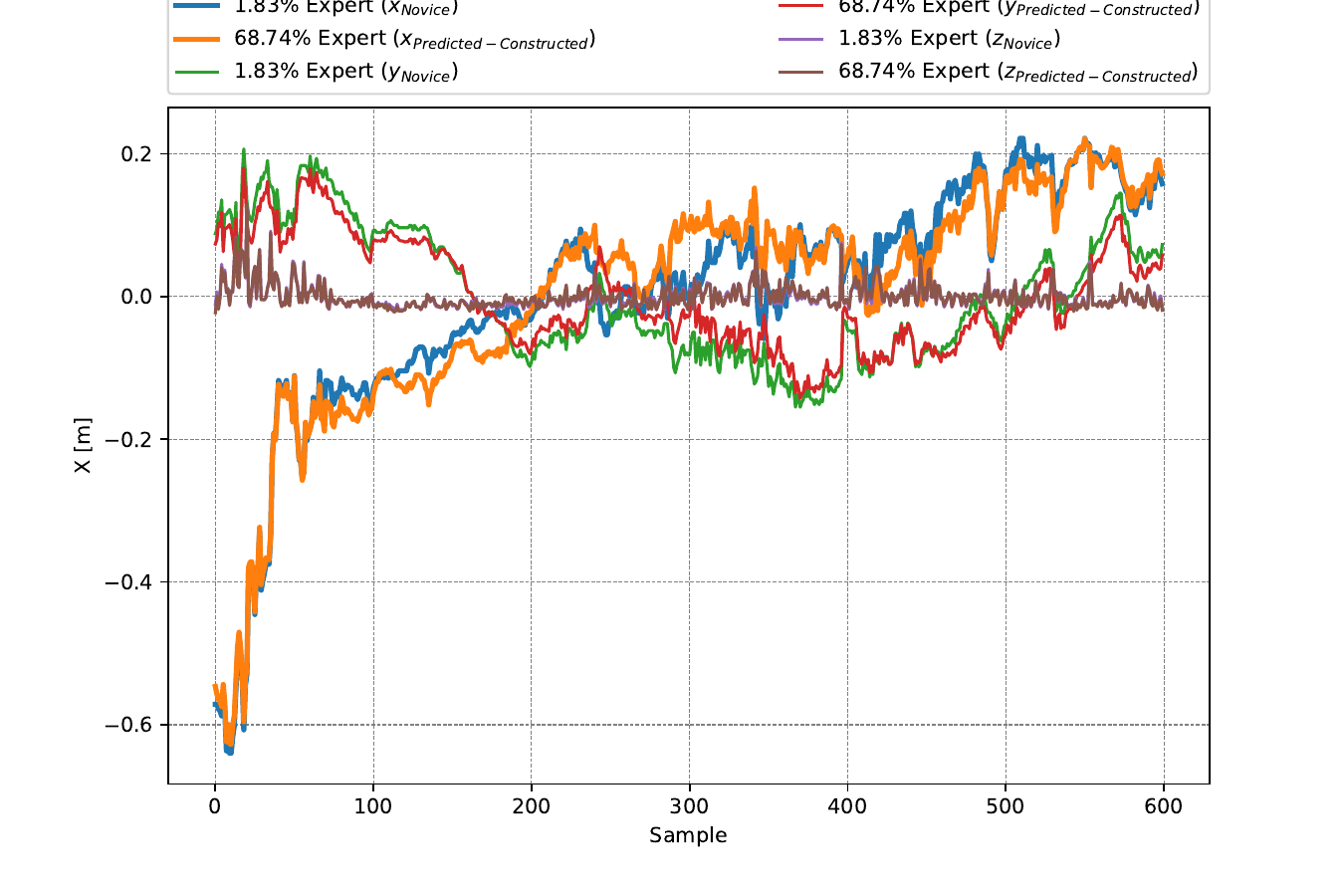}
		\label{fig5}
	}
	\subfigure[ ]
	{
		\includegraphics[width=0.45\columnwidth]{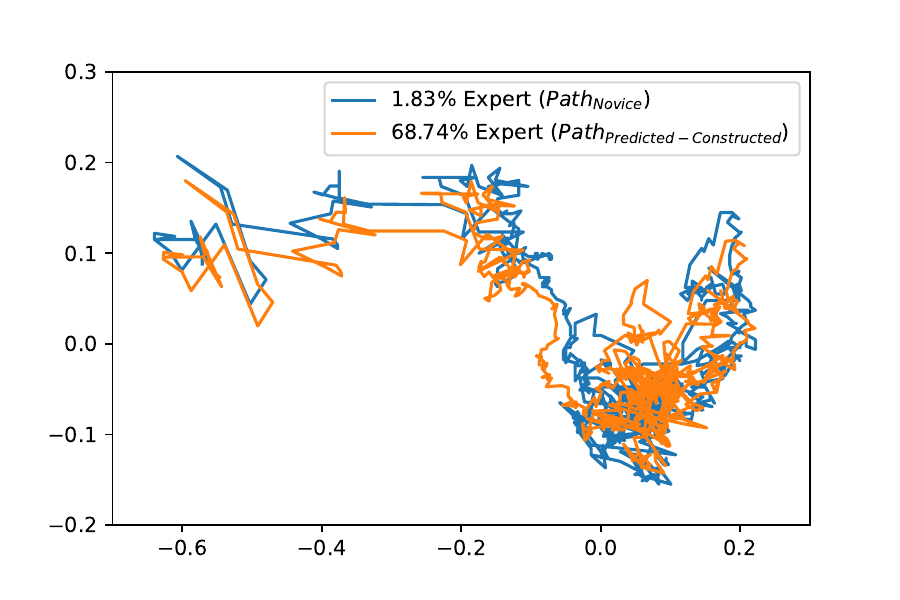}
		\label{fig6}
	}
	\caption{
		Comparison of the base path (novice) and the predicted-constructed path (expert) for four different surgical procedures performed by an expert surgeon.}
	\label{Psss122}
\end{figure}

In the final experiment given in this section, the information related to a novice surgeon is considered, and the skills of four different surgical procedures performed by an expert surgeon are transferred to him. With the same coefficients, it can be seen that the proposed path becomes 55\% more expert on average. \autoref{Psss122} contains the graphs associated with this experiment.
\begin{table}[!t]
\caption{Skill enhancement of the novice surgeon using two different expert paths.}
    \label{tab33}
    \centering
    \begin{tabular}{|l|c|c|}
\hline Indicators and expert paths & ID1 (4.35/5) & ID17 (5/5) \\
\hline Total Skill Level Improvement & $13.20$ & $20.33$ \\
\hline Total Predictability Improvement & $16.34$ & $18.87$ \\
\hline Enhancement in Tremor Reduction & $2.33$ & $3.44$ \\
\hline Enhancement in Noise Cancelation & $1.32$ & $2.40$ \\
\hline
\end{tabular}
\end{table}


\subsection{JIGSAWS Dataset}
\begin{figure}[h!]
	\centering
	\subfigure[$\frac{b}{a}=1.5$]{
		\includegraphics[width=0.45\columnwidth]{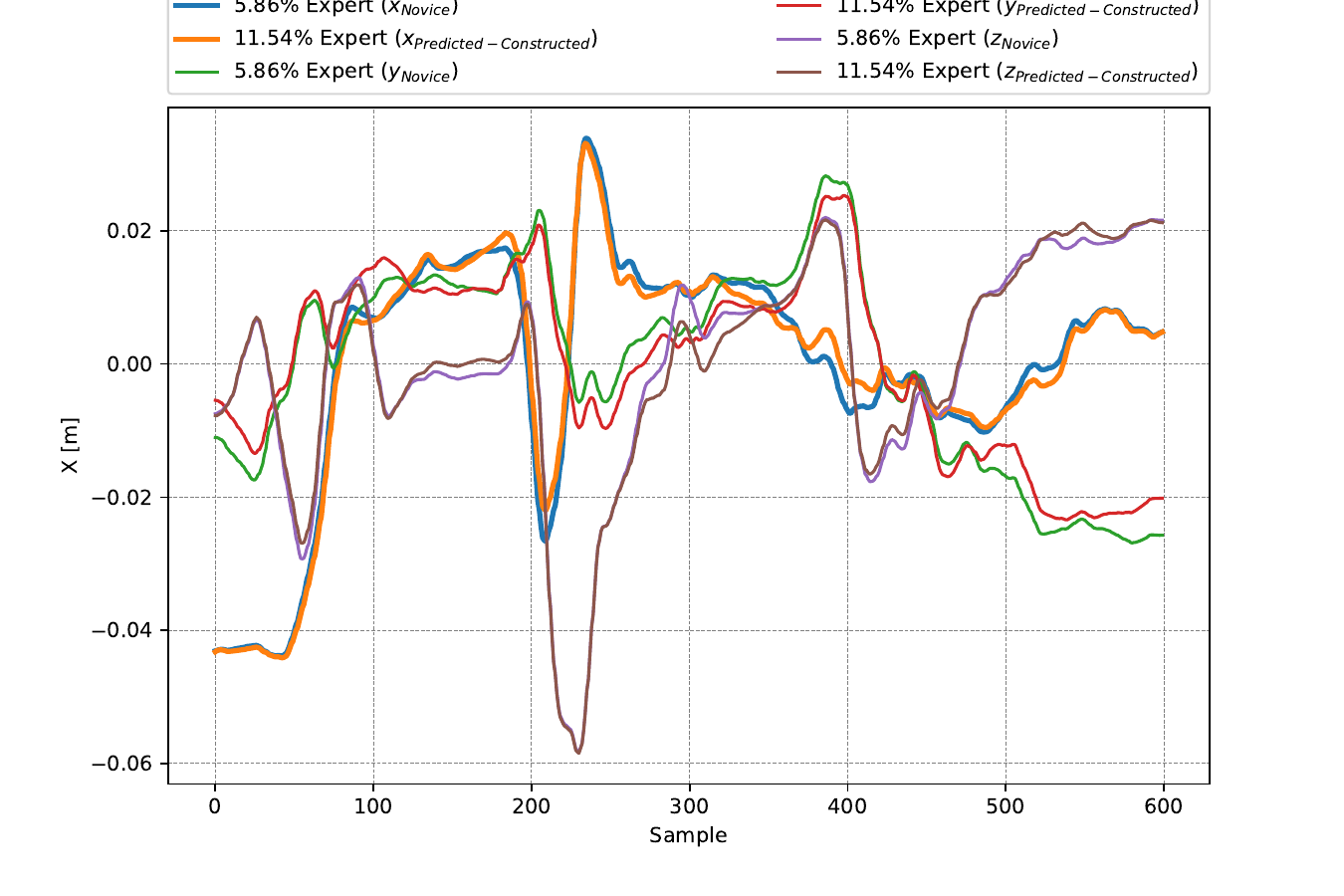}
		\label{fig55}
	}
	\subfigure[$\frac{b}{a}=1.5$]
	{
		\includegraphics[width=0.45\columnwidth]{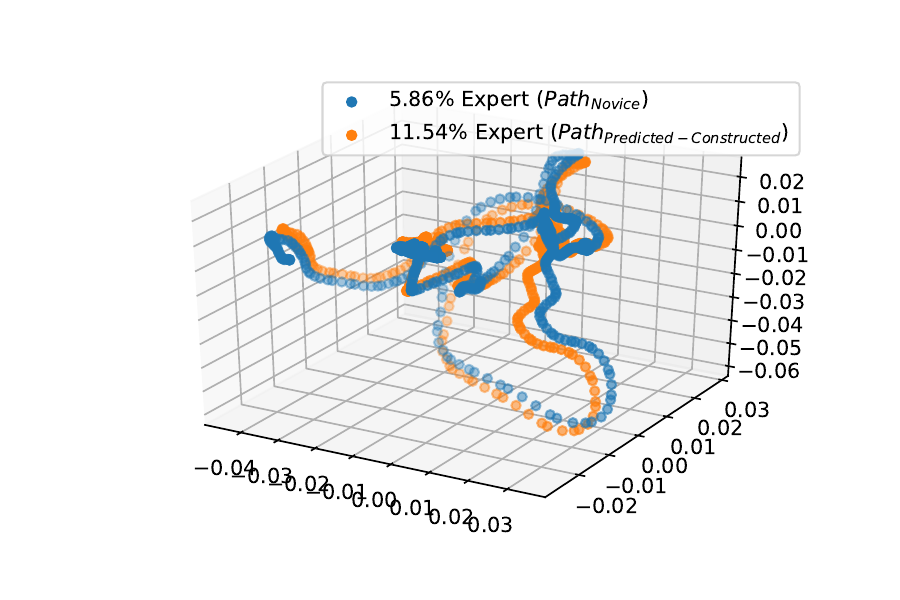}
		\label{fig56}
	}
 \\
 \subfigure[$\frac{b}{a}=5$]{
		\includegraphics[width=0.45\columnwidth]{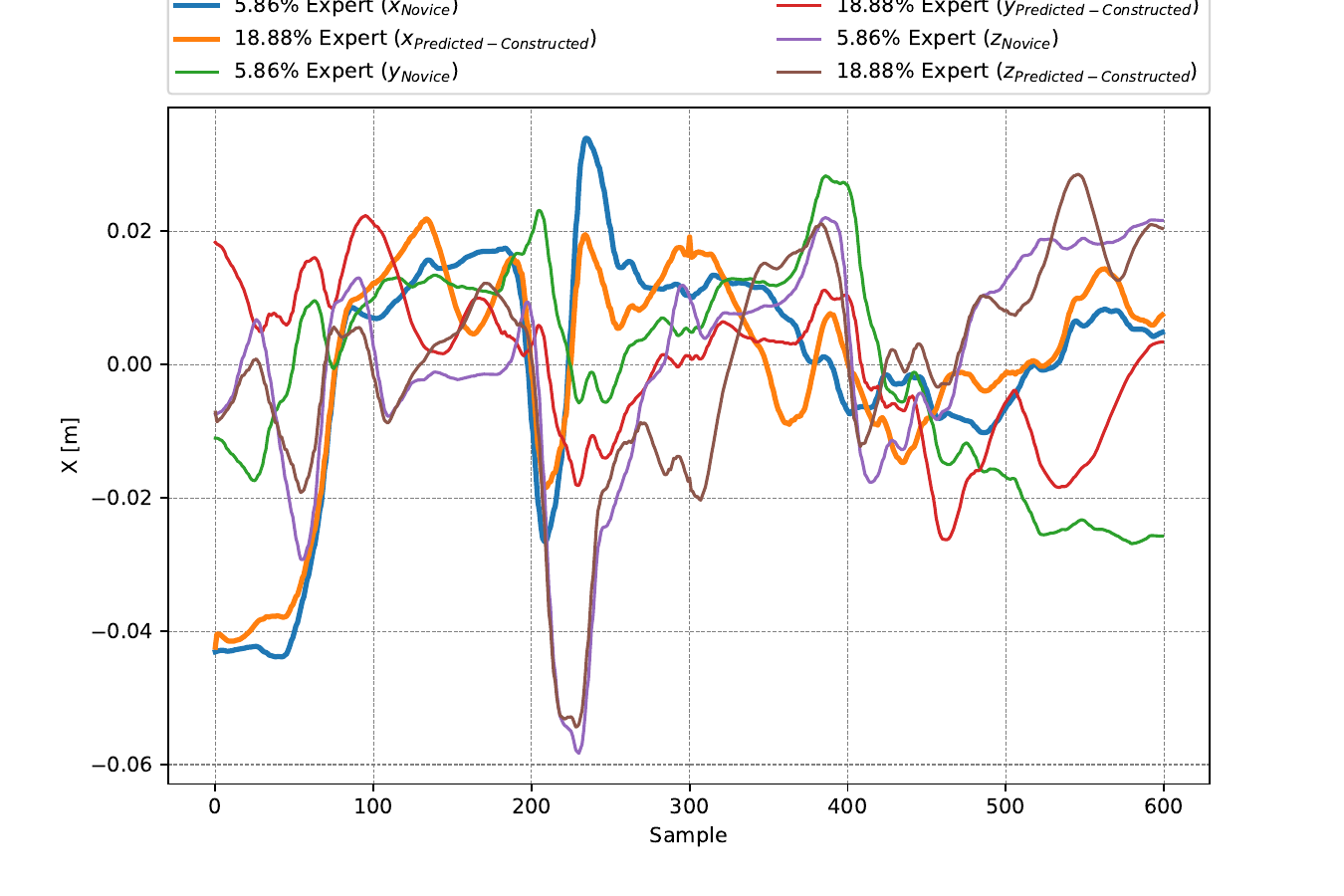}
		\label{fig525}
	}
	\subfigure[$\frac{b}{a}=5$]
	{
		\includegraphics[width=0.45\columnwidth]{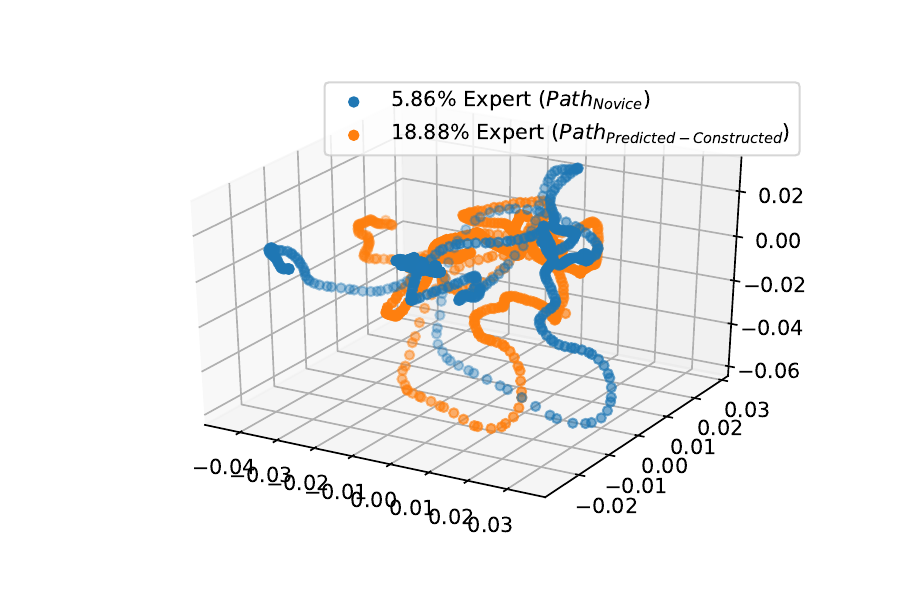}
		\label{fig526}
	}
	\caption{
		Comparison of the base path (novice) and the predicted-constructed path (expert).}
	\label{P12222}
\end{figure}

Since the JIGSAWS dataset contains information on x, y, and z, the skill transfer structure is applied to these variables. As shown in \autoref{P12222}, when the significance coefficient of the loss function linked to skill improves, the suggested movement path becomes increasingly expert. \autoref{tab1} gives further information regarding the degree of change in the rookie surgeon's skill level when employing the proposed  path. Using the proposed platform, the revised path of the inexperienced surgeon will feature more expert traits and fewer surprising motions and tremors.
\begin{table}[ht!]
\caption{Skill enhancement of the novice surgeon using the predicted-constructed path.}
    \label{tab1}
    \centering
    \begin{tabular}{|l|c|c|}
\hline Indicators and Ratios of Coefficients & $\frac{\boldsymbol{b}}{\boldsymbol{a}}=\mathbf{1 . 5}$ & $\frac{\boldsymbol{b}}{\boldsymbol{a}}=\mathbf{5}$ \\
\hline Total Skill Level Improvement & $5.67$ & $13.01$ \\
\hline Total Predictability Improvement & $8.95$ & $5.96$ \\
\hline Enhancement in Tremor Reduction & $5.91$ & $18.44$ \\
\hline Enhancement in Noise Cancelation & $0.39$ & $3.40$ \\
\hline
\end{tabular}  
\end{table}

%
\section{Conclusion}\label{s3}
In this study, a CNN-FFT network pre-trained with motion data from the JIGSAWS and ARAS-Farabi datasets is presented. The network generates a feature space, referred to as a reference model when expert surgeon data is used as input to the network.
When the network is given the movement path of the novice surgeon, a proposed movement path is constructed using two loss functions and an optimization procedure. The suggested path not only takes into account the trainee surgeon's objective but also enhances their skill level.
The amount of improvement was demonstrated by employing the newly developed convolutional network and some concrete indicators.
It has been demonstrated that the inclusion of a valuable AI agent for skill transfer in the surgical training cycle can be accomplished through the utilization of this platform.
It was also demonstrated that as the skill level of the reference model increases, there is a greater likelihood of enhancing the movement patterns of trainee surgeons.
This research offers an advantage over previous studies by preserving the surgeon's aim in capsulorhexis surgery, although the precise nature of that aim is not specified.
This approach is suitable for a wide range of surgical training applications because it emphasizes both general and specific (capsulorhexis) surgical abilities. By using a robotic haptic device and impedance control structure, trainee surgeons can sense and learn the suggested path applied to their hands.



\ifCLASSOPTIONcompsoc
  \section*{Acknowledgments}
\else
  \section*{Acknowledgment}
\fi

NIMAD-INSF Grant...

\ifCLASSOPTIONcaptionsoff
  \newpage
\fi

\end{document}